\pdfoutput=1

\documentclass[11pt,a4paper]{article}
\PassOptionsToPackage{hyphens}{url}
\usepackage{times,latexsym}
\usepackage{url}
\usepackage[T1]{fontenc}

\usepackage[acceptedWithA]{tacl2018v2}

\usepackage{amsmath}
\usepackage{amssymb}
\usepackage{mathtools}
\usepackage{arydshln}
\usepackage{booktabs,colortbl}
\usepackage{tabularx}
\usepackage{multirow}
\usepackage{tikz}
\usepackage{pgfplots}
\usepackage[normalem]{ulem}
\usepackage{xcolor}
\usepackage{dashbox}%
\usepackage{textcomp}
\usepackage{tcolorbox}
\usepackage{adjustbox}
\usepackage{lipsum}
\usepackage[inline]{enumitem}
\usepackage{menukeys}
\usepackage[letterspace=-20]{microtype}
\usepackage{pmboxdraw}
\usepackage{sansmath}
\usepackage{titlesec}

\pgfplotsset{compat=1.13}

\definecolor{c0}{cmyk}{1,0.3968,0,0.2588} 
\definecolor{c1}{cmyk}{0,0.6175,0.8848,0.1490} 
\definecolor{c2}{cmyk}{0.1127,0.6690,0,0.4431} 
\definecolor{c3}{cmyk}{0.3081,0,0.7209,0.3255} 
\definecolor{c4}{cmyk}{0.6765,0.2017,0,0.0667} 
\definecolor{c5}{cmyk}{0,0.8765,0.7099,0.3647} 

\definecolor{c0alt}{RGB}{15,158,251} 

\definecolor{darkgrey}{RGB}{149,149,149}
\definecolor{decentgrey}{RGB}{242,242,242}

\usetikzlibrary{calc,fit,positioning,arrows,arrows.meta,backgrounds,decorations.pathreplacing}
\usepgfplotslibrary{fillbetween}

\usetikzlibrary{external}

\pgfdeclarelayer{bg}
\pgfsetlayers{bg,main}

\newtcbox{\hlprimary}{on line,colback=c0!10,colframe=white,size=fbox,arc=3pt, box align=base,before upper=\strut, top=-2pt, bottom=-4pt, left=-1pt, right=-1pt, boxrule=0pt}
\newtcbox{\hlprimarytab}{on line, box align=base, colback=c0!10,colframe=white,size=fbox,arc=3pt, before upper=\strut, top=-2pt, bottom=-4pt, left=-2pt, right=-2pt, boxrule=0pt}
\newtcbox{\hlsecondary}{on line,colback=c1!10,colframe=white,size=fbox,arc=3pt, box align=base,before upper=\strut, top=-2pt, bottom=-4pt, left=-1pt, right=-1pt, boxrule=0pt}
\newtcbox{\hlsecondarytab}{on line, box align=base, colback=c1!10,colframe=white,size=fbox,arc=3pt, before upper=\strut, top=-2pt, bottom=-4pt, left=-2pt, right=-2pt, boxrule=0pt}
\newtcolorbox{hlmultiline}{on line,colback=decentgrey!75,colframe=white,size=fbox,arc=3pt, box align=base, top=0pt, bottom=2pt, boxrule=0pt, before=\adjustbox{valign=c}\bgroup, after=\egroup, before upper=\strut}

\newcolumntype{Y}{>{\centering\arraybackslash}X}
\newcolumntype{Z}{>{\raggedleft\arraybackslash}X}

\newcommand{\pet}{\textsc{Pet}}

\newcommand{\ipet}{i\pet{}}

\newcommand{\bt}{\fontseries{b}\selectfont}
\newcommand{\negphantom}[1]{\settowidth{\dimen0}{#1}\hspace*{-\dimen0}}

\titleclass{\question}{straight}[\subsection]
\newcounter{question}
\titleformat{\question}
{\normalfont\normalsize\bfseries}{}{0em}{Q\thequestion:~}
\titlespacing*{\question}{0pt}{3.25ex plus 1ex minus .2ex}{1.5ex plus .2ex}

\title{True Few-Shot Learning with Prompts -- A Real-World Perspective}

\author{Timo Schick \and Hinrich Sch\"{u}tze\\[0.5em]
Center for Information and Language Processing (CIS), LMU Munich, Germany\\[0.5em]
{\tt schickt@cis.lmu.de}, {\tt inquiries@cislmu.org}
}

\date{}

\newcounter{notecounter}
\newcommand{\enotesoff}{\long\gdef\enote##1##2{}}

\enotesoff

\begin{document}
\maketitle
\begin{abstract}

Prompt-based approaches are strong at few-shot learning.
However,
\citet{perez2021true} have recently cast doubt on their
performance because they had difficulty getting good
results in a ``true'' few-shot setting in which
prompts and hyperparameters cannot be tuned on a dev set.
In view of this, we conduct an extensive study of \pet{}, a method that
combines textual instructions with example-based finetuning. We show that, if
correctly configured, \pet{} performs strongly in a true
few-shot setting, i.e., without a dev set. 
Crucial
for this strong performance  is \pet's ability
to intelligently handle multiple prompts.  We then put our findings to a
real-world test by running \pet{} on RAFT, a benchmark of
tasks taken directly from realistic NLP applications for which no 
labeled dev or test sets are available.
\pet{} achieves a new state of the art on RAFT and performs close
to non-expert humans for 7 out of 11 tasks. These results
demonstrate that prompt-based learners like \pet{} excel at
true few-shot learning and
underpin our belief that learning from instructions
will
play an important role on the path towards human-like
few-shot learning capabilities.

\end{abstract}

\section{Introduction}

With pretrained language models (LMs) getting ever larger \citep{radford2018language,raffel2019exploring,brown2020language,fedus2021switch}, \emph{instruction-based learning} has emerged as a powerful method for few-shot text classification \citep[e.g.,][]{jiang2019know,schick2020exploiting,schick2020just,brown2020language,wei2021finetuned,sanh2021multitask}. The key idea is to give an LM access to descriptive names for all possible outputs and to short prompts explaining the task to be solved. In settings where at most a few dozen examples are available, this simple idea leads to substantial improvements over various baselines \citep{schick2020exploiting,schick2020just,gao2020making,tam2021improving}.

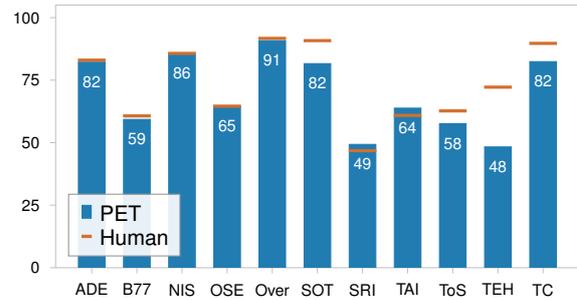
\begin{figure}
	\centering
	\begin{tikzpicture}
	\begin{axis}[
	ybar,
	symbolic x coords={ade, b77, nisr, ose, or, sot, sri, tai, tos, teh, tc},
	xticklabels={ADE, B77,NIS,OSE,Over,SOT,SRI,TAI,ToS,TEH,TC}, 
	xtick=data,
	ytick={0, 25, 50, 75, 100},
	bar width=0.35cm,
	axis line style={decentgrey!80!black},
	major tick style={decentgrey!80!black},
	xtick pos=left,
	ytick pos=left,
	ylabel near ticks,
	xlabel near ticks,
	tick align=outside,
	enlarge x limits=0.08,
	ymin=0,
	ymax=105,
	major tick length=0.075cm,
	width = 1.1\linewidth,
	height = 0.205\textheight,
	log ticks with fixed point,
	every node near coord/.append style={font=\sansmath\sffamily\tiny, inner sep=5pt, color=white, anchor=north, /pgf/number format/.cd, fixed, fixed zerofill, precision=0, /tikz/.cd},
	x tick label style={font=\sansmath\sffamily\tiny},
	y tick label style={font=\sansmath\sffamily\tiny, /pgf/number format/.cd, fixed, fixed zerofill, precision=0, /tikz/.cd},
	/tikz/mstyle/.style={draw,c1, very thick,text=black},
	/tikz/mleft/.style={xshift={-(.5*\pgfplotbarwidth+0.2pt)}},
	/tikz/mline/.style={insert path={--+({\pgfplotbarwidth+0.4pt},0pt)}},
	]
	
	
	\addplot[c0!75, fill=c0!75, nodes near coords] coordinates {(ade,82.2) (b77,59.3) (nisr,85.7) (ose,64.6) (or,90.8) (sot,81.6) (sri,49.3) (tai,63.8) (tos,57.6) (teh,48.3) (tc,82.4)};
	
	\path[mstyle]([mleft]axis cs:ade,83.0)[mline]node[right]{};
	\path[mstyle]([mleft]axis cs:b77,60.7)[mline]node[right]{};
	\path[mstyle]([mleft]axis cs:nisr,85.7)[mline]node[right]{};
	\path[mstyle]([mleft]axis cs:ose,64.6)[mline]node[right]{};
	\path[mstyle]([mleft]axis cs:or,91.7)[mline]node[right]{};
	\path[mstyle]([mleft]axis cs:sot,90.8)[mline]node[right]{};
	\path[mstyle]([mleft]axis cs:sri,46.8)[mline]node[right]{};
	\path[mstyle]([mleft]axis cs:tai,60.9)[mline]node[right]{};
	\path[mstyle]([mleft]axis cs:tos,62.7)[mline]node[right]{};
	\path[mstyle]([mleft]axis cs:teh,72.2)[mline]node[right]{};
	\path[mstyle]([mleft]axis cs:tc,89.7)[mline]node[right]{};
	
	\node[anchor=north west, xshift=0.1cm, inner sep=0](legend-pet) at (axis cs:ade,25) {\scriptsize\sffamily PET};
	\node[left=0.1cm of legend-pet, draw=c0!75, fill=c0!75, inner sep=0, yshift=0.02cm, minimum width=0.15cm, minimum height=0.17cm](legend-symbol-pet){};
	
	\node[below=0.1cm of legend-pet.south west, anchor=north west, inner sep=0](legend-human) {\scriptsize\sffamily Human};
	\node[left=0.1cm of legend-human, draw=c1, fill=c1, inner sep=0, yshift=0.01cm, minimum width=0.15cm, minimum height=0.03cm](legend-symbol-human){};
	\node[draw=darkgrey, fit=(legend-pet)(legend-symbol-pet)(legend-human)(legend-symbol-human), inner sep=4pt, fill=white, fill opacity=0.9]{};
	
	\node[anchor=north west, xshift=0.1cm, inner sep=0](legend-pet) at (axis cs:ade,25) {\scriptsize\sffamily PET};
	\node[left=0.1cm of legend-pet, draw=c0!75, fill=c0!75, inner sep=0, yshift=0.02cm, minimum width=0.12cm, minimum height=0.17cm](legend-symbol-pet){};
	
	\node[below=0.1cm of legend-pet.south west, anchor=north west, inner sep=0](legend-human) {\scriptsize\sffamily Human};
	\node[left=0.1cm of legend-human, draw=c1, fill=c1, inner sep=0, yshift=0.01cm, minimum width=0.12cm, minimum height=0.03cm](legend-symbol-human){};
	
	\end{axis}
	\end{tikzpicture}
	\caption{\pet{} achieves near-human performance for
        7 out of 11 tasks
of the RAFT benchmark \citep{alex2021raft}, for which labeled development and
        test sets are not available.
This
demonstrates that prompt-based learners like \pet{}, if
        correctly configured, excel at
true few-shot learning, i.e.,
without any tuning of instructions or hyperparameters on a development set.}
	\label{fig:intro-raft}
\end{figure}

However, recent work has questioned the strong few-shot performance of instruction-based approaches, arguing in particular that the considered settings are often not \emph{true} few-shot settings \citep{perez2021true,logan2021cutting} mainly for two reasons: For one, some approaches \citep[e.g.,][]{xie2019unsupervised,zhang2019pegasus,chen2020mixtext,tam2021improving} make use of large development sets to optimize hyperparameters. Beyond that, it is argued that manually designed instructions require manual tuning on development sets to achieve strong performance \citep{perez2021true,logan2021cutting}. Indeed, performance can vary largely -- and in mostly unpredictable ways -- across different instructions \citep{jiang2019know,schick2020exploiting}; this issue even persists after finetuning a model on hundreds of instructions \citep{sanh2021multitask}. Even separate from this problem, the need for human involvement is generally seen as a huge drawback of manually designed instructions \citep{shin2020autoprompt,lester2021power}. Thus, several recent works abandon them in favor of automatically generated prompts \citep{shin2020autoprompt,gao2020making,hambardzumyan-etal-2021-warp,li2021prefixtuning,lester2021power}.

Contrary to this trend, we argue that when correctly configured, prompt-based approaches achieve strong performance even in true few-shot settings and that there is no problem in using manually designed instructions per se. On the opposite, such instructions are often relatively easy to specify if one is familiar with the task to be solved, they provide an intuitive interface to convey task-specific knowledge%
, and if properly used, they consistently improve model performance in few-shot settings. To provide empirical support for these claims, we revisit \pet{} \citep{schick2020exploiting} -- a method for combining instructions with example-based finetuning whose key feature is that it allows users to specify \emph{multiple} instructions for a single task -- and thoroughly examine its performance with human-made instructions in true few-shot settings. In order to simulate a real-world scenario as best as possible, we proceed in two steps: First, we conduct an extensive study of \pet{} using three English academic datasets to analyze its ability to perform true few-shot learning in a controlled environment and to derive best practices regarding the choice of instructions and other hyperparameters. We then put our findings to the test and evaluate \pet{} on a large variety of different real-world tasks from the RAFT benchmark \citep{alex2021raft}, for which no labeled development or test sets are available, enforcing a \emph{true} few-shot setting \citep{perez2021true}. On average, \pet{} clearly outperforms all baselines on this dataset and comes surprisingly close to the performance of non-expert humans (see Figure~\ref{fig:intro-raft}), demonstrating that instruction-based learning can successfully be applied to real-world tasks in true few-shot settings.

In summary, the main contributions of this work are as follows:

\begin{itemize}
	\item We investigate the performance of \pet{} for various models, tasks and training set sizes, its ability to cope with different instructions and its robustness to hyperparameter choices in true few-shot settings.
	\item We show how \pet{} can be used when no unlabeled data is available and propose a variant for efficient classification in scenarios with many different classes.
	\item We apply \pet{} to RAFT \citep{alex2021raft}, a benchmark of real-world tasks where it obtains a new state of the art and achieves near-human performance for 7 out of 11 tasks in true few-shot settings. 
\end{itemize}

\section{Related Work}

As a precursor to instruction-based learning, some works have investigated ways of informing classifiers about the meaning of different output classes both for text \citep{chang2008importance,veeranna2016using,zhou2019zero} and image classification \citep{norouzi2014zeroshot,romera2015embarrassingly}; actually providing instructions in the form of short prompts was first proposed by \citet{radford2018language}. This idea has since been applied to solve a wide range of different NLP tasks without any task-specific training data \citep{puri2019zeroshot,opitz2019argumentative,davison-etal-2019-commonsense,schick2021selfdiagnosis,schick2021generating,wei2021finetuned,sanh2021multitask}. While most approaches rephrase tasks as a language modeling problem, some use prompts to reformulate them as different tasks for which large amounts of training data are available \citep{levy-etal-2017-zero,mccann2018natural,yin-etal-2019-benchmarking,sun2021nspbert,sainz2021label}. Instruction-based learning has also been used in few-shot settings; popular variants include \emph{in-context learning}, where the model's parameters are fixed and examples are provided as additional context \citep{brown2020language,lu2021fantastically,kumar2021reordering,min2021noisy}, \emph{finetuning} the entire model \citep{schick2020exploiting,schick2020just,gao2020making,tam2021improving}, and \emph{prompt tuning}, where only the instruction itself is optimized \citep{shin2020autoprompt,hambardzumyan-etal-2021-warp,li2021prefixtuning,lester2021power}.

\begin{figure*}
	\tikzset{
		every node/.style={
			font=\sffamily\footnotesize,
		},
		tbox/.style={
			draw=decentgrey!80!black, rounded corners, inner sep=5pt, outer sep=3pt, 
		},
		header/.style={
			inner sep=0,
			outer sep=2pt,
		},
		pbox/.style={
			tbox, minimum width=6.3cm, text width=6.3cm
		},
		vbox/.style={
			tbox, minimum width=3.5cm, text width=3.5cm
		},
		arrow/.style={
			draw=decentgrey!80!black,-,>=latex
		},
	}
	\centering
	\begin{tikzpicture}
	
	\node[pbox](p1){\textcolor{c0}{I really enjoyed this movie.} \textcolor{c1}{Question: Is this a positive movie review?} \textcolor{c1}{Answer: {\scriptsize [MASK]}.}};
	\node[pbox, above=0.05cm of p1.north west, anchor=south west](p0){\textcolor{c0}{I really enjoyed this movie.} \textcolor{c1}{It was {\scriptsize [MASK]}.}};
	\node[pbox, below=0.05cm of p1.south west, anchor=north west, align=left](p2){\textcolor{c1}{(\,{\scriptsize [MASK]}\,)} \textcolor{c0}{I really enjoyed this movie.}};
	
	\node[inner sep=0, outer sep=0, fit=(p0)(p1)(p2)](p-wrapper){};
	\node[tbox, left=0.5cm of p-wrapper](input){\textcolor{c0}{I really enjoyed this movie.}};
	
	\node[vbox, right=0.5cm of p0](v0){\textcolor{c2}{ $+$ = good\negphantom{good}\phantom{\lsstyle positive} $-$ = bad}};		
	\node[vbox, right=0.5cm of p1](v1){\textcolor{c2}{ $+$ = Yes\negphantom{Yes}\phantom{\lsstyle positive} $-$ = No}};
	\node[vbox, right=0.5cm of p2](v2){\textcolor{c2}{ $+$ = {\lsstyle positive} $-$ = \lsstyle negative}};		
	
	\node[header, above=0cm of input](input-header){\scalebox{0.9}{\textcolor{c0}{\textbf{INPUT}}}};		
	\node[header, above=0cm of p0](pattern-header){\scalebox{0.9}{\textcolor{c1}{\textbf{PATTERNS}}}};		
	\node[header, above=0cm of v0](pattern-verbalizer){\scalebox{0.9}{\textcolor{c2}{\textbf{VERBALIZERS}}}};
	
	\path[] ([yshift=0.15cm]input.east) edge[arrow, out=0, in=180] (p0.west);
	\path[] (input.east) edge[arrow] (p1.west);
	\path[] ([yshift=-0.15cm]input.east) edge[arrow, out=0, in=180] (p2.west);
	
	\path[] (p0) edge[arrow] (v0);
	\path[] (p1) edge[arrow] (v1);
	\path[] (p2) edge[arrow] (v2);
	
	\end{tikzpicture}
	\caption{Different choices of \textcolor{c1}{patterns} and corresponding \textcolor{c2}{verbalizers} for classifying movie reviews as \emph{positive} ($+$) or \emph{negative} ($-$). The \textcolor{c0}{input} is first converted into a cloze question using the pattern; classification is done by computing the output whose verbalization is the most likely substitute for the mask according to the MLM.}
	\label{fig:patterns-and-verbalizers}
\end{figure*}
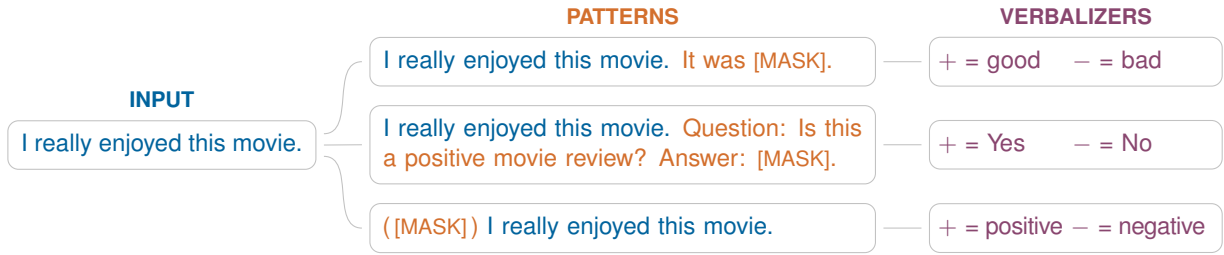

Several works investigating the limitations and drawbacks of instruction-based few-shot approaches find that current LMs are mostly unable to understand complex instructions that go beyond short prompts or simple questions \citep{efrat2020turking,weller2020learning,webson2021promptbased} and that they are highly sensitive to the exact wording of the instructions provided \citep{jiang2019know,schick2020exploiting,elazar2021measuring}. In a similar vein, \citet{perez2021true} and \citet{logan2021cutting} argue that prior work overestimates few-shot performance as manual prompt tuning is required to achieve good performance. Accordingly, some works attempt to obtain either prompts \citep{shin2020autoprompt,gao2020making,li2021prefixtuning,lester2021power} or meaningful names for output classes \citep{schick-etal-2020-automatically,gao2020making} without any human involvement.

Finally, many benchmarks have been proposed for comparing few-shot approaches in a standardized way \citep[e.g.,][]{mishra2021crosstask,bragg2021flex,xu2021fewclue,ye2021crossfit,alex2021raft}. As our focus is on the real-world applicability of few-shot methods, we evaluate \pet{} on the RAFT benchmark \citep{alex2021raft}, which measures performance in applied settings.

\section{Pattern-Exploiting Training}

We briefly review \emph{pattern-exploiting training} (\pet{}) \citep{schick2020exploiting,schick2020just}, the method we use for instruction-based text classification. 
At its core, \pet{} combines textual instructions with regular finetuning using labeled examples. To that end, users must specify one or more \emph{patterns} which convert an input example $x$ into a cloze question so that it can readily be processed by a masked language model (MLM) \citep{devlin2018bert}.\footnote{We use the term \emph{prompt} to refer to a short sequence of tokens that typically contains some form of instruction; the term \emph{pattern} is used to denote the function that adds a prompt to an input.} These patterns can take on very different forms; some examples are shown in Figure~\ref{fig:patterns-and-verbalizers}. In addition, users must inform the model about the meaning of all output classes; this is done with a \emph{verbalizer} that assigns a natural language expression to each output $y$ (see Figure~\ref{fig:patterns-and-verbalizers}, right). We refer to the combination of a pattern and verbalizer as a \emph{pattern-verbalizer pair} (PVP).

Given a single PVP, let $p(y \mid x)$ be the probability that an MLM assigns to $y$'s verbalization in the cloze question obtained by applying the pattern to $x$, normalized over all $y$. The MLM is finetuned on labeled examples $(x, y)$ by minimizing the cross-entropy loss between $p(y \mid x)$ and $y$.

If a user specifies \emph{multiple} PVPs, individual models are trained for each pair. Similar to knowledge distillation \citep{hinton2015distilling}, they are then used to annotate unlabeled examples for training a final classifier with a regular sequence classification head \citep{devlin2018bert}. We use the \emph{weighted} variant of \pet{} without auxiliary language modeling; see \citet{schick2020exploiting} for details.

\paragraph{Multi-Token Verbalizers} When using encoder-only MLMs \citep{devlin2018bert,liu2019roberta,lan2019albert}, one limitation of \pet{} is that the verbalization of each output class must correspond to a single token. \citet{schick2020just} propose a fix for this that uses multiple mask tokens, but their approach is very inefficient. 
As discussed by \citet{schick2021fewshot}, an alternative solution is to instead use an encoder-decoder LM \citep{lewis2019bart,raffel2019exploring}. In Section~\ref{sec:raft}, we propose yet another solution that allows us to stick with encoder-only MLMs while being much more efficient than the approach of \citet{schick2020just}.

\paragraph{Iterative \pet{}} We also experiment with \ipet{} \citep{schick2020exploiting}, an iterative variant of \pet{} that employs self-training \citep[e.g.,][]{scudder1965probability,yarowsky-1995-unsupervised,brin1999extracting,mcclosky-etal-2006-effective}
to train several generations of models on datasets of increasing size. To this end, an ensemble of MLMs is trained as in regular \pet{} and then used to assign labels to unlabeled examples; a new ensemble is trained on the so-obtained data. This process is repeated for multiple iterations, where the number of annotated examples is increased in each iteration. We refer to \citet{schick2020exploiting} for more details.

\section{True Few-Shot Learning with \pet{}}
\label{sec:true-fsl}

We conduct a variety of experiments to answer six important research questions (Q\ref{q1}--Q\ref{q6}) regarding the extent to which true few-shot learning is possible with \pet{}. Beyond that, the purpose of our experiments is to establish a set of best practices for our real-world experiments on RAFT \citep{alex2021raft}. Before discussing individual experiments, we describe the underlying setup.

\paragraph{Tasks and Datasets} While they are heavily used in prior work \citep[e.g.,][]{brown2020language,schick2020just,logan2021cutting,webson2021promptbased}, we decide against tasks and datasets from the GLUE \citep{wang-etal-2018-glue} and SuperGLUE benchmarks \citep{wang2019superglue} as they are very different from what we expect to see in real-world applications. Instead, we experiment with AG's News, Yelp Reviews Full Star and Yahoo Questions \citep{zhang2015character} as these datasets represent classification tasks in three different domains that resemble real-world settings. Further, it is relatively easy to come up with a variety of instructions for each of these tasks, making it more straightforward to experiment with a large number of different patterns.

We consider settings with $n = 10$ and $n = 100$ training examples. For each $n$, we generate five different training sets per task by randomly sampling examples from the original training sets while ensuring that the number of examples is about the same for each possible output class. In addition, for both $n= 10$ and $n = 100$, we sample 1,000 unlabeled examples from the original training sets. We repeat all of our experiments for all five training sets and, by default, report average performance.

\paragraph{PVPs} We manually write a total of 23 patterns per task, all of which can be categorized into one of the following groups:\footnote{The full set of PVPs can be found in Appendix~\ref{appendix:default-pvps}.}

\begin{itemize}
	\item {\textsc{Null}}: Following \citet{logan2021cutting}, these patterns simply insert a mask token.
	\item {\textsc{Punc}}: Similar to some patterns of \citet{schick2020exploiting}, these patterns only add punctuation characters and a mask token.
	\item {\textsc{Prompts}}: Patterns in this group add short prompts -- typically consisting of no more than three words -- to the input, similar to \citet{radford2018language} and \citet{schick2020exploiting}.
	\item {\textsc{Q\&A}}: These patterns rephrase the task as a question $q$ and append
	\[ \small\textsf{Question: \emph{q} Answer: {\footnotesize [MASK]}.}\] to the input, similar to \citet{brown2020language} and \citet{schick2021selfdiagnosis}.
\end{itemize} 
For all patterns, we use only a single verbalizer which we adopt from \citet{schick2020exploiting}. While varying the verbalizer may also lead to interesting insights, finding a large amount of reasonable verbalizers is challenging for some tasks as there is often a single, \emph{natural} choice (e.g., the actual category names for AG's News and Yahoo Questions).

\paragraph{Hyperparameters} We consider a setting similar to that of \citet{schick2020exploiting,schick2020just} and, unless otherwise specified, use the default settings of the \pet{} library.\footnote{See \url{https://github.com/timoschick/pet}} As our experiments require training hundreds of models, we make a few changes to reduce environmental impact \citep{strubell-etal-2019-energy} and computational cost: We use the base variant of RoBERTa \citep{liu2019roberta} as underlying LM, we train only one model per PVP, and we reduce the training steps for all individual models and the final classifier to 100 and 1,000, respectively.

\paragraph{Monitoring} Finetuning pretrained LMs can be unstable on small datasets \citep{devlin2018bert,dodge2020finetuning}, sometimes leading to very poor performance. Luckily, we can detect such finetuning issues to some extent even without a labeled test set using the following two checks: 
\begin{itemize}
	\item \textsc{Train Set Underfitting}: We check for training runs that result in less than 50\% accuracy on the training set. As finetuning on up to 100 examples typically leads to perfect predictions on the training set, this is a clear indicator of a failed finetuning run.
	\item \textsc{Constant Predictions}: Another strong indicator of unsuccessful training is when the finetuned model predicts the same output class for all inputs; we check this both on the training set and on the unlabeled set.
\end{itemize}
Whenever one of these events occurs, we restart training using a different seed for initializing all random number generators.

\question{How can we find \emph{the} best pattern -- or do we even need to?}
\label{q1}

\pgfplotsset{
	bplot/.style={
		axis line style={decentgrey!80!black},
		ytick distance={0.05},
		xtick={0,1,2,3,4,5,6,7,8,9,10,11,12,13,14,15,16,17,18,19,20,21,22,24,26},
		major tick style={decentgrey!80!black},
		xtick pos=left,
		ytick pos=left,
		ylabel near ticks,
		xlabel near ticks,
		tick align=outside,
		enlarge x limits=0.02,
		title style={yshift=-1.5ex},
		enlarge y limits=0.05,
		major tick length=0.075cm,
		width = 0.54\linewidth,
		height = 0.15\textheight,
		log ticks with fixed point,
		x tick label style={font=\sansmath\sffamily\tiny},
		y tick label style={font=\sansmath\sffamily\tiny, /pgf/number format/.cd, fixed, fixed zerofill, precision=2, /tikz/.cd},
	},
}

\begin{figure*}
\centering
\begin{tikzpicture}
\begin{axis}[bplot, ytick distance={0.05}, xticklabels={\color{c1}2\vphantom{0},\color{c1}0\vphantom{0},\color{c2}5\vphantom{0},\color{c1}1\vphantom{0},\color{c2}7\vphantom{0},\color{c2}6\vphantom{0},\color{c3}10\vphantom{0},\color{c2}8\vphantom{0},\color{c3}14\vphantom{0},\color{c2}3\vphantom{0},\color{c3}15\vphantom{0},\color{c3}13\vphantom{0},\color{c4}21\vphantom{0},\color{c3}9\vphantom{0},\color{c3}16\vphantom{0},\color{c3}11\vphantom{0},\color{c4}22\vphantom{0},\color{c2}4\vphantom{0},\color{c3}12\vphantom{0},\color{c3}17\vphantom{0},\color{c4}19\vphantom{0},\color{c4}20\vphantom{0},\color{c3}18\vphantom{0},\color{black}PET\vphantom{0},\color{black}iPET\vphantom{0}}, title={\scriptsize\sffamily AG's News ($n = 10$)}]
\addplot[decentgrey!80!black, dashed] coordinates {(0,0.7695389016018308)(26,0.7695389016018308)};
\addplot[mark=*, mark size=1pt, c1, mark options={solid}, dotted] coordinates {(0,0.705)(0,0.673)(0,0.638)(0,0.646)(0,0.750)};
\addplot[mark=*, mark size=1.8pt, c1, thick, mark options={solid}, only marks] coordinates {(0,0.683)};
\addplot[mark=*, mark size=1pt, c1, mark options={solid}, dotted] coordinates {(1,0.726)(1,0.629)(1,0.712)(1,0.670)(1,0.690)};
\addplot[mark=*, mark size=1.8pt, c1, thick, mark options={solid}, only marks] coordinates {(1,0.685)};
\addplot[mark=*, mark size=1pt, c2, mark options={solid}, dotted] coordinates {(2,0.754)(2,0.647)(2,0.665)(2,0.735)(2,0.757)};
\addplot[mark=*, mark size=1.8pt, c2, thick, mark options={solid}, only marks] coordinates {(2,0.711)};
\addplot[mark=*, mark size=1pt, c1, mark options={solid}, dotted] coordinates {(3,0.759)(3,0.708)(3,0.680)(3,0.732)(3,0.682)};
\addplot[mark=*, mark size=1.8pt, c1, thick, mark options={solid}, only marks] coordinates {(3,0.712)};
\addplot[mark=*, mark size=1pt, c2, mark options={solid}, dotted] coordinates {(4,0.791)(4,0.727)(4,0.653)(4,0.753)(4,0.731)};
\addplot[mark=*, mark size=1.8pt, c2, thick, mark options={solid}, only marks] coordinates {(4,0.731)};
\addplot[mark=*, mark size=1pt, c2, mark options={solid}, dotted] coordinates {(5,0.727)(5,0.803)(5,0.698)(5,0.733)(5,0.797)};
\addplot[mark=*, mark size=1.8pt, c2, thick, mark options={solid}, only marks] coordinates {(5,0.752)};
\addplot[mark=*, mark size=1pt, c3, mark options={solid}, dotted] coordinates {(6,0.801)(6,0.778)(6,0.715)(6,0.721)(6,0.776)};
\addplot[mark=*, mark size=1.8pt, c3, thick, mark options={solid}, only marks] coordinates {(6,0.758)};
\addplot[mark=*, mark size=1pt, c2, mark options={solid}, dotted] coordinates {(7,0.766)(7,0.757)(7,0.748)(7,0.747)(7,0.787)};
\addplot[mark=*, mark size=1.8pt, c2, thick, mark options={solid}, only marks] coordinates {(7,0.761)};
\addplot[mark=*, mark size=1pt, c3, mark options={solid}, dotted] coordinates {(8,0.776)(8,0.790)(8,0.727)(8,0.708)(8,0.841)};
\addplot[mark=*, mark size=1.8pt, c3, thick, mark options={solid}, only marks] coordinates {(8,0.768)};
\addplot[mark=*, mark size=1pt, c2, mark options={solid}, dotted] coordinates {(9,0.769)(9,0.773)(9,0.778)(9,0.794)(9,0.794)};
\addplot[mark=*, mark size=1.8pt, c2, thick, mark options={solid}, only marks] coordinates {(9,0.782)};
\addplot[mark=*, mark size=1pt, c3, mark options={solid}, dotted] coordinates {(10,0.792)(10,0.798)(10,0.768)(10,0.772)(10,0.793)};
\addplot[mark=*, mark size=1.8pt, c3, thick, mark options={solid}, only marks] coordinates {(10,0.785)};
\addplot[mark=*, mark size=1pt, c3, mark options={solid}, dotted] coordinates {(11,0.801)(11,0.799)(11,0.797)(11,0.709)(11,0.819)};
\addplot[mark=*, mark size=1.8pt, c3, thick, mark options={solid}, only marks] coordinates {(11,0.785)};
\addplot[mark=*, mark size=1pt, c4, mark options={solid}, dotted] coordinates {(12,0.766)(12,0.768)(12,0.804)(12,0.786)(12,0.806)};
\addplot[mark=*, mark size=1.8pt, c4, thick, mark options={solid}, only marks] coordinates {(12,0.786)};
\addplot[mark=*, mark size=1pt, c3, mark options={solid}, dotted] coordinates {(13,0.800)(13,0.817)(13,0.763)(13,0.787)(13,0.780)};
\addplot[mark=*, mark size=1.8pt, c3, thick, mark options={solid}, only marks] coordinates {(13,0.790)};
\addplot[mark=*, mark size=1pt, c3, mark options={solid}, dotted] coordinates {(14,0.797)(14,0.806)(14,0.776)(14,0.782)(14,0.800)};
\addplot[mark=*, mark size=1.8pt, c3, thick, mark options={solid}, only marks] coordinates {(14,0.792)};
\addplot[mark=*, mark size=1pt, c3, mark options={solid}, dotted] coordinates {(15,0.796)(15,0.799)(15,0.768)(15,0.774)(15,0.832)};
\addplot[mark=*, mark size=1.8pt, c3, thick, mark options={solid}, only marks] coordinates {(15,0.794)};
\addplot[mark=*, mark size=1pt, c4, mark options={solid}, dotted] coordinates {(16,0.779)(16,0.808)(16,0.814)(16,0.799)(16,0.780)};
\addplot[mark=*, mark size=1.8pt, c4, thick, mark options={solid}, only marks] coordinates {(16,0.796)};
\addplot[mark=*, mark size=1pt, c2, mark options={solid}, dotted] coordinates {(17,0.790)(17,0.828)(17,0.770)(17,0.789)(17,0.819)};
\addplot[mark=*, mark size=1.8pt, c2, thick, mark options={solid}, only marks] coordinates {(17,0.799)};
\addplot[mark=*, mark size=1pt, c3, mark options={solid}, dotted] coordinates {(18,0.801)(18,0.786)(18,0.792)(18,0.789)(18,0.841)};
\addplot[mark=*, mark size=1.8pt, c3, thick, mark options={solid}, only marks] coordinates {(18,0.801)};
\addplot[mark=*, mark size=1pt, c3, mark options={solid}, dotted] coordinates {(19,0.776)(19,0.818)(19,0.788)(19,0.827)(19,0.801)};
\addplot[mark=*, mark size=1.8pt, c3, thick, mark options={solid}, only marks] coordinates {(19,0.802)};
\addplot[mark=*, mark size=1pt, c4, mark options={solid}, dotted] coordinates {(20,0.760)(20,0.819)(20,0.808)(20,0.800)(20,0.834)};
\addplot[mark=*, mark size=1.8pt, c4, thick, mark options={solid}, only marks] coordinates {(20,0.804)};
\addplot[mark=*, mark size=1pt, c4, mark options={solid}, dotted] coordinates {(21,0.782)(21,0.837)(21,0.803)(21,0.802)(21,0.819)};
\addplot[mark=*, mark size=1.8pt, c4, thick, mark options={solid}, only marks] coordinates {(21,0.808)};
\addplot[mark=*, mark size=1pt, c3, mark options={solid}, dotted] coordinates {(22,0.801)(22,0.831)(22,0.820)(22,0.791)(22,0.828)};
\addplot[mark=*, mark size=1.8pt, c3, thick, mark options={solid}, only marks] coordinates {(22,0.814)};
\addplot[mark=*, mark size=1pt, black, mark options={solid}, dotted] coordinates {(24,0.817)(24,0.844)(24,0.803)(24,0.821)(24,0.830)};
\addplot[mark=*, mark size=1.8pt, black, thick, mark options={solid}, only marks] coordinates {(24,0.823)};
\addplot[mark=*, mark size=1pt, black, mark options={solid}, dotted] coordinates {(26,0.831)(26,0.869)(26,0.847)(26,0.840)(26,0.839)};
\addplot[mark=*, mark size=1.8pt, black, thick, mark options={solid}, only marks] coordinates {(26,0.845)};
\end{axis}
\end{tikzpicture}%
$\;$%
\begin{tikzpicture}
\begin{axis}[bplot, ytick distance={0.02}, xticklabels={\color{c1}2\vphantom{0},\color{c1}0\vphantom{0},\color{c1}1\vphantom{0},\color{c3}11\vphantom{0},\color{c2}7\vphantom{0},\color{c3}9\vphantom{0},\color{c3}15\vphantom{0},\color{c3}14\vphantom{0},\color{c2}5\vphantom{0},\color{c3}13\vphantom{0},\color{c3}16\vphantom{0},\color{c2}3\vphantom{0},\color{c3}12\vphantom{0},\color{c3}10\vphantom{0},\color{c2}6\vphantom{0},\color{c4}21\vphantom{0},\color{c3}18\vphantom{0},\color{c4}20\vphantom{0},\color{c2}8\vphantom{0},\color{c4}22\vphantom{0},\color{c2}4\vphantom{0},\color{c4}19\vphantom{0},\color{c3}17\vphantom{0},\color{black}PET\vphantom{0},\color{black}iPET\vphantom{0}}, title={\scriptsize\sffamily AG's News ($n = 100$)}]
\addplot[decentgrey!80!black, dashed] coordinates {(0,0.8690400457665903)(26,0.8690400457665903)};
\addplot[mark=*, mark size=1pt, c1, mark options={solid}, dotted] coordinates {(0,0.839)(0,0.864)(0,0.807)(0,0.860)(0,0.842)};
\addplot[mark=*, mark size=1.8pt, c1, thick, mark options={solid}, only marks] coordinates {(0,0.842)};
\addplot[mark=*, mark size=1pt, c1, mark options={solid}, dotted] coordinates {(1,0.857)(1,0.854)(1,0.831)(1,0.850)(1,0.850)};
\addplot[mark=*, mark size=1.8pt, c1, thick, mark options={solid}, only marks] coordinates {(1,0.848)};
\addplot[mark=*, mark size=1pt, c1, mark options={solid}, dotted] coordinates {(2,0.856)(2,0.871)(2,0.852)(2,0.856)(2,0.879)};
\addplot[mark=*, mark size=1.8pt, c1, thick, mark options={solid}, only marks] coordinates {(2,0.863)};
\addplot[mark=*, mark size=1pt, c3, mark options={solid}, dotted] coordinates {(3,0.864)(3,0.869)(3,0.868)(3,0.863)(3,0.861)};
\addplot[mark=*, mark size=1.8pt, c3, thick, mark options={solid}, only marks] coordinates {(3,0.865)};
\addplot[mark=*, mark size=1pt, c2, mark options={solid}, dotted] coordinates {(4,0.872)(4,0.859)(4,0.860)(4,0.863)(4,0.876)};
\addplot[mark=*, mark size=1.8pt, c2, thick, mark options={solid}, only marks] coordinates {(4,0.866)};
\addplot[mark=*, mark size=1pt, c3, mark options={solid}, dotted] coordinates {(5,0.867)(5,0.875)(5,0.866)(5,0.853)(5,0.877)};
\addplot[mark=*, mark size=1.8pt, c3, thick, mark options={solid}, only marks] coordinates {(5,0.868)};
\addplot[mark=*, mark size=1pt, c3, mark options={solid}, dotted] coordinates {(6,0.865)(6,0.879)(6,0.856)(6,0.863)(6,0.876)};
\addplot[mark=*, mark size=1.8pt, c3, thick, mark options={solid}, only marks] coordinates {(6,0.868)};
\addplot[mark=*, mark size=1pt, c3, mark options={solid}, dotted] coordinates {(7,0.865)(7,0.876)(7,0.868)(7,0.856)(7,0.877)};
\addplot[mark=*, mark size=1.8pt, c3, thick, mark options={solid}, only marks] coordinates {(7,0.868)};
\addplot[mark=*, mark size=1pt, c2, mark options={solid}, dotted] coordinates {(8,0.863)(8,0.877)(8,0.855)(8,0.878)(8,0.872)};
\addplot[mark=*, mark size=1.8pt, c2, thick, mark options={solid}, only marks] coordinates {(8,0.869)};
\addplot[mark=*, mark size=1pt, c3, mark options={solid}, dotted] coordinates {(9,0.877)(9,0.873)(9,0.845)(9,0.869)(9,0.881)};
\addplot[mark=*, mark size=1.8pt, c3, thick, mark options={solid}, only marks] coordinates {(9,0.869)};
\addplot[mark=*, mark size=1pt, c3, mark options={solid}, dotted] coordinates {(10,0.858)(10,0.872)(10,0.853)(10,0.884)(10,0.883)};
\addplot[mark=*, mark size=1.8pt, c3, thick, mark options={solid}, only marks] coordinates {(10,0.870)};
\addplot[mark=*, mark size=1pt, c2, mark options={solid}, dotted] coordinates {(11,0.871)(11,0.877)(11,0.856)(11,0.870)(11,0.879)};
\addplot[mark=*, mark size=1.8pt, c2, thick, mark options={solid}, only marks] coordinates {(11,0.871)};
\addplot[mark=*, mark size=1pt, c3, mark options={solid}, dotted] coordinates {(12,0.873)(12,0.872)(12,0.865)(12,0.877)(12,0.870)};
\addplot[mark=*, mark size=1.8pt, c3, thick, mark options={solid}, only marks] coordinates {(12,0.871)};
\addplot[mark=*, mark size=1pt, c3, mark options={solid}, dotted] coordinates {(13,0.873)(13,0.873)(13,0.868)(13,0.879)(13,0.873)};
\addplot[mark=*, mark size=1.8pt, c3, thick, mark options={solid}, only marks] coordinates {(13,0.873)};
\addplot[mark=*, mark size=1pt, c2, mark options={solid}, dotted] coordinates {(14,0.869)(14,0.884)(14,0.867)(14,0.868)(14,0.881)};
\addplot[mark=*, mark size=1.8pt, c2, thick, mark options={solid}, only marks] coordinates {(14,0.874)};
\addplot[mark=*, mark size=1pt, c4, mark options={solid}, dotted] coordinates {(15,0.876)(15,0.881)(15,0.867)(15,0.879)(15,0.868)};
\addplot[mark=*, mark size=1.8pt, c4, thick, mark options={solid}, only marks] coordinates {(15,0.874)};
\addplot[mark=*, mark size=1pt, c3, mark options={solid}, dotted] coordinates {(16,0.874)(16,0.885)(16,0.874)(16,0.871)(16,0.868)};
\addplot[mark=*, mark size=1.8pt, c3, thick, mark options={solid}, only marks] coordinates {(16,0.874)};
\addplot[mark=*, mark size=1pt, c4, mark options={solid}, dotted] coordinates {(17,0.874)(17,0.876)(17,0.869)(17,0.878)(17,0.876)};
\addplot[mark=*, mark size=1.8pt, c4, thick, mark options={solid}, only marks] coordinates {(17,0.875)};
\addplot[mark=*, mark size=1pt, c2, mark options={solid}, dotted] coordinates {(18,0.876)(18,0.880)(18,0.861)(18,0.884)(18,0.874)};
\addplot[mark=*, mark size=1.8pt, c2, thick, mark options={solid}, only marks] coordinates {(18,0.875)};
\addplot[mark=*, mark size=1pt, c4, mark options={solid}, dotted] coordinates {(19,0.886)(19,0.874)(19,0.860)(19,0.881)(19,0.875)};
\addplot[mark=*, mark size=1.8pt, c4, thick, mark options={solid}, only marks] coordinates {(19,0.875)};
\addplot[mark=*, mark size=1pt, c2, mark options={solid}, dotted] coordinates {(20,0.883)(20,0.881)(20,0.868)(20,0.868)(20,0.879)};
\addplot[mark=*, mark size=1.8pt, c2, thick, mark options={solid}, only marks] coordinates {(20,0.876)};
\addplot[mark=*, mark size=1pt, c4, mark options={solid}, dotted] coordinates {(21,0.879)(21,0.884)(21,0.860)(21,0.881)(21,0.875)};
\addplot[mark=*, mark size=1.8pt, c4, thick, mark options={solid}, only marks] coordinates {(21,0.876)};
\addplot[mark=*, mark size=1pt, c3, mark options={solid}, dotted] coordinates {(22,0.868)(22,0.892)(22,0.867)(22,0.879)(22,0.885)};
\addplot[mark=*, mark size=1.8pt, c3, thick, mark options={solid}, only marks] coordinates {(22,0.878)};
\addplot[mark=*, mark size=1pt, black, mark options={solid}, dotted] coordinates {(24,0.884)(24,0.879)(24,0.875)(24,0.886)(24,0.885)};
\addplot[mark=*, mark size=1.8pt, black, thick, mark options={solid}, only marks] coordinates {(24,0.882)};
\addplot[mark=*, mark size=1pt, black, mark options={solid}, dotted] coordinates {(26,0.879)(26,0.877)(26,0.877)(26,0.878)(26,0.876)};
\addplot[mark=*, mark size=1.8pt, black, thick, mark options={solid}, only marks] coordinates {(26,0.877)};
\end{axis}
\end{tikzpicture}%
\vskip3pt
\begin{tikzpicture}
\begin{axis}[bplot, ytick distance={0.05}, xticklabels={\color{c1}0\vphantom{0},\color{c1}1\vphantom{0},\color{c2}2\vphantom{0},\color{c2}3\vphantom{0},\color{c2}6\vphantom{0},\color{c3}13\vphantom{0},\color{c2}4\vphantom{0},\color{c3}10\vphantom{0},\color{c2}5\vphantom{0},\color{c3}7\vphantom{0},\color{c3}16\vphantom{0},\color{c3}17\vphantom{0},\color{c3}14\vphantom{0},\color{c3}15\vphantom{0},\color{c3}8\vphantom{0},\color{c3}9\vphantom{0},\color{c4}20\vphantom{0},\color{c3}18\vphantom{0},\color{c4}21\vphantom{0},\color{c3}11\vphantom{0},\color{c3}12\vphantom{0},\color{c4}19\vphantom{0},\color{c4}22\vphantom{0},\color{black}PET\vphantom{0},\color{black}iPET\vphantom{0}}, title={\scriptsize\sffamily Yelp Full ($n = 10$)}]
\addplot[decentgrey!80!black, dashed] coordinates {(0,0.43025043478260866)(26,0.43025043478260866)};
\addplot[mark=*, mark size=1pt, c1, mark options={solid}, dotted] coordinates {(0,0.382)(0,0.352)(0,0.387)(0,0.331)(0,0.355)};
\addplot[mark=*, mark size=1.8pt, c1, thick, mark options={solid}, only marks] coordinates {(0,0.361)};
\addplot[mark=*, mark size=1pt, c1, mark options={solid}, dotted] coordinates {(1,0.397)(1,0.406)(1,0.354)(1,0.377)(1,0.389)};
\addplot[mark=*, mark size=1.8pt, c1, thick, mark options={solid}, only marks] coordinates {(1,0.384)};
\addplot[mark=*, mark size=1pt, c2, mark options={solid}, dotted] coordinates {(2,0.424)(2,0.405)(2,0.379)(2,0.354)(2,0.405)};
\addplot[mark=*, mark size=1.8pt, c2, thick, mark options={solid}, only marks] coordinates {(2,0.393)};
\addplot[mark=*, mark size=1pt, c2, mark options={solid}, dotted] coordinates {(3,0.428)(3,0.400)(3,0.393)(3,0.352)(3,0.401)};
\addplot[mark=*, mark size=1.8pt, c2, thick, mark options={solid}, only marks] coordinates {(3,0.395)};
\addplot[mark=*, mark size=1pt, c2, mark options={solid}, dotted] coordinates {(4,0.421)(4,0.408)(4,0.388)(4,0.433)(4,0.445)};
\addplot[mark=*, mark size=1.8pt, c2, thick, mark options={solid}, only marks] coordinates {(4,0.419)};
\addplot[mark=*, mark size=1pt, c3, mark options={solid}, dotted] coordinates {(5,0.430)(5,0.437)(5,0.412)(5,0.434)(5,0.404)};
\addplot[mark=*, mark size=1.8pt, c3, thick, mark options={solid}, only marks] coordinates {(5,0.424)};
\addplot[mark=*, mark size=1pt, c2, mark options={solid}, dotted] coordinates {(6,0.446)(6,0.435)(6,0.438)(6,0.380)(6,0.421)};
\addplot[mark=*, mark size=1.8pt, c2, thick, mark options={solid}, only marks] coordinates {(6,0.424)};
\addplot[mark=*, mark size=1pt, c3, mark options={solid}, dotted] coordinates {(7,0.436)(7,0.434)(7,0.415)(7,0.418)(7,0.422)};
\addplot[mark=*, mark size=1.8pt, c3, thick, mark options={solid}, only marks] coordinates {(7,0.425)};
\addplot[mark=*, mark size=1pt, c2, mark options={solid}, dotted] coordinates {(8,0.435)(8,0.432)(8,0.439)(8,0.404)(8,0.434)};
\addplot[mark=*, mark size=1.8pt, c2, thick, mark options={solid}, only marks] coordinates {(8,0.429)};
\addplot[mark=*, mark size=1pt, c3, mark options={solid}, dotted] coordinates {(9,0.437)(9,0.441)(9,0.397)(9,0.419)(9,0.451)};
\addplot[mark=*, mark size=1.8pt, c3, thick, mark options={solid}, only marks] coordinates {(9,0.429)};
\addplot[mark=*, mark size=1pt, c3, mark options={solid}, dotted] coordinates {(10,0.458)(10,0.456)(10,0.397)(10,0.446)(10,0.413)};
\addplot[mark=*, mark size=1.8pt, c3, thick, mark options={solid}, only marks] coordinates {(10,0.434)};
\addplot[mark=*, mark size=1pt, c3, mark options={solid}, dotted] coordinates {(11,0.473)(11,0.441)(11,0.384)(11,0.454)(11,0.419)};
\addplot[mark=*, mark size=1.8pt, c3, thick, mark options={solid}, only marks] coordinates {(11,0.434)};
\addplot[mark=*, mark size=1pt, c3, mark options={solid}, dotted] coordinates {(12,0.459)(12,0.443)(12,0.402)(12,0.426)(12,0.452)};
\addplot[mark=*, mark size=1.8pt, c3, thick, mark options={solid}, only marks] coordinates {(12,0.436)};
\addplot[mark=*, mark size=1pt, c3, mark options={solid}, dotted] coordinates {(13,0.475)(13,0.458)(13,0.399)(13,0.431)(13,0.429)};
\addplot[mark=*, mark size=1.8pt, c3, thick, mark options={solid}, only marks] coordinates {(13,0.439)};
\addplot[mark=*, mark size=1pt, c3, mark options={solid}, dotted] coordinates {(14,0.471)(14,0.462)(14,0.404)(14,0.444)(14,0.427)};
\addplot[mark=*, mark size=1.8pt, c3, thick, mark options={solid}, only marks] coordinates {(14,0.442)};
\addplot[mark=*, mark size=1pt, c3, mark options={solid}, dotted] coordinates {(15,0.451)(15,0.469)(15,0.452)(15,0.441)(15,0.439)};
\addplot[mark=*, mark size=1.8pt, c3, thick, mark options={solid}, only marks] coordinates {(15,0.450)};
\addplot[mark=*, mark size=1pt, c4, mark options={solid}, dotted] coordinates {(16,0.472)(16,0.464)(16,0.426)(16,0.447)(16,0.444)};
\addplot[mark=*, mark size=1.8pt, c4, thick, mark options={solid}, only marks] coordinates {(16,0.451)};
\addplot[mark=*, mark size=1pt, c3, mark options={solid}, dotted] coordinates {(17,0.471)(17,0.452)(17,0.425)(17,0.438)(17,0.471)};
\addplot[mark=*, mark size=1.8pt, c3, thick, mark options={solid}, only marks] coordinates {(17,0.451)};
\addplot[mark=*, mark size=1pt, c4, mark options={solid}, dotted] coordinates {(18,0.467)(18,0.475)(18,0.422)(18,0.437)(18,0.467)};
\addplot[mark=*, mark size=1.8pt, c4, thick, mark options={solid}, only marks] coordinates {(18,0.453)};
\addplot[mark=*, mark size=1pt, c3, mark options={solid}, dotted] coordinates {(19,0.486)(19,0.441)(19,0.399)(19,0.462)(19,0.481)};
\addplot[mark=*, mark size=1.8pt, c3, thick, mark options={solid}, only marks] coordinates {(19,0.454)};
\addplot[mark=*, mark size=1pt, c3, mark options={solid}, dotted] coordinates {(20,0.471)(20,0.461)(20,0.419)(20,0.447)(20,0.476)};
\addplot[mark=*, mark size=1.8pt, c3, thick, mark options={solid}, only marks] coordinates {(20,0.455)};
\addplot[mark=*, mark size=1pt, c4, mark options={solid}, dotted] coordinates {(21,0.483)(21,0.499)(21,0.413)(21,0.422)(21,0.463)};
\addplot[mark=*, mark size=1.8pt, c4, thick, mark options={solid}, only marks] coordinates {(21,0.456)};
\addplot[mark=*, mark size=1pt, c4, mark options={solid}, dotted] coordinates {(22,0.483)(22,0.481)(22,0.421)(22,0.444)(22,0.458)};
\addplot[mark=*, mark size=1.8pt, c4, thick, mark options={solid}, only marks] coordinates {(22,0.457)};
\addplot[mark=*, mark size=1pt, black, mark options={solid}, dotted] coordinates {(24,0.502)(24,0.489)(24,0.416)(24,0.446)(24,0.479)};
\addplot[mark=*, mark size=1.8pt, black, thick, mark options={solid}, only marks] coordinates {(24,0.466)};
\addplot[mark=*, mark size=1pt, black, mark options={solid}, dotted] coordinates {(26,0.543)(26,0.538)(26,0.457)(26,0.464)(26,0.514)};
\addplot[mark=*, mark size=1.8pt, black, thick, mark options={solid}, only marks] coordinates {(26,0.503)};
\end{axis}
\end{tikzpicture}%
$\;$%
\begin{tikzpicture}
\begin{axis}[bplot, ytick distance={0.02}, xticklabels={\color{c1}1\vphantom{0},\color{c1}0\vphantom{0},\color{c2}3\vphantom{0},\color{c3}7\vphantom{0},\color{c2}2\vphantom{0},\color{c2}4\vphantom{0},\color{c2}6\vphantom{0},\color{c3}10\vphantom{0},\color{c3}13\vphantom{0},\color{c3}16\vphantom{0},\color{c3}8\vphantom{0},\color{c2}5\vphantom{0},\color{c4}22\vphantom{0},\color{c3}17\vphantom{0},\color{c3}18\vphantom{0},\color{c4}21\vphantom{0},\color{c3}15\vphantom{0},\color{c3}14\vphantom{0},\color{c4}19\vphantom{0},\color{c3}12\vphantom{0},\color{c4}20\vphantom{0},\color{c3}9\vphantom{0},\color{c3}11\vphantom{0},\color{black}PET\vphantom{0},\color{black}iPET\vphantom{0}}, title={\scriptsize\sffamily Yelp Full ($n = 100$)}]
\addplot[decentgrey!80!black, dashed] coordinates {(0,0.5376765217391303)(26,0.5376765217391303)};
\addplot[mark=*, mark size=1pt, c1, mark options={solid}, dotted] coordinates {(0,0.484)(0,0.506)(0,0.471)(0,0.508)(0,0.473)};
\addplot[mark=*, mark size=1.8pt, c1, thick, mark options={solid}, only marks] coordinates {(0,0.488)};
\addplot[mark=*, mark size=1pt, c1, mark options={solid}, dotted] coordinates {(1,0.488)(1,0.491)(1,0.497)(1,0.475)(1,0.506)};
\addplot[mark=*, mark size=1.8pt, c1, thick, mark options={solid}, only marks] coordinates {(1,0.491)};
\addplot[mark=*, mark size=1pt, c2, mark options={solid}, dotted] coordinates {(2,0.516)(2,0.534)(2,0.501)(2,0.529)(2,0.527)};
\addplot[mark=*, mark size=1.8pt, c2, thick, mark options={solid}, only marks] coordinates {(2,0.521)};
\addplot[mark=*, mark size=1pt, c3, mark options={solid}, dotted] coordinates {(3,0.497)(3,0.544)(3,0.520)(3,0.549)(3,0.500)};
\addplot[mark=*, mark size=1.8pt, c3, thick, mark options={solid}, only marks] coordinates {(3,0.522)};
\addplot[mark=*, mark size=1pt, c2, mark options={solid}, dotted] coordinates {(4,0.514)(4,0.536)(4,0.521)(4,0.537)(4,0.516)};
\addplot[mark=*, mark size=1.8pt, c2, thick, mark options={solid}, only marks] coordinates {(4,0.525)};
\addplot[mark=*, mark size=1pt, c2, mark options={solid}, dotted] coordinates {(5,0.531)(5,0.525)(5,0.520)(5,0.534)(5,0.517)};
\addplot[mark=*, mark size=1.8pt, c2, thick, mark options={solid}, only marks] coordinates {(5,0.525)};
\addplot[mark=*, mark size=1pt, c2, mark options={solid}, dotted] coordinates {(6,0.540)(6,0.527)(6,0.513)(6,0.536)(6,0.519)};
\addplot[mark=*, mark size=1.8pt, c2, thick, mark options={solid}, only marks] coordinates {(6,0.527)};
\addplot[mark=*, mark size=1pt, c3, mark options={solid}, dotted] coordinates {(7,0.511)(7,0.549)(7,0.540)(7,0.535)(7,0.525)};
\addplot[mark=*, mark size=1.8pt, c3, thick, mark options={solid}, only marks] coordinates {(7,0.532)};
\addplot[mark=*, mark size=1pt, c3, mark options={solid}, dotted] coordinates {(8,0.512)(8,0.562)(8,0.541)(8,0.554)(8,0.513)};
\addplot[mark=*, mark size=1.8pt, c3, thick, mark options={solid}, only marks] coordinates {(8,0.536)};
\addplot[mark=*, mark size=1pt, c3, mark options={solid}, dotted] coordinates {(9,0.528)(9,0.562)(9,0.541)(9,0.546)(9,0.515)};
\addplot[mark=*, mark size=1.8pt, c3, thick, mark options={solid}, only marks] coordinates {(9,0.539)};
\addplot[mark=*, mark size=1pt, c3, mark options={solid}, dotted] coordinates {(10,0.546)(10,0.553)(10,0.555)(10,0.557)(10,0.514)};
\addplot[mark=*, mark size=1.8pt, c3, thick, mark options={solid}, only marks] coordinates {(10,0.545)};
\addplot[mark=*, mark size=1pt, c2, mark options={solid}, dotted] coordinates {(11,0.529)(11,0.560)(11,0.526)(11,0.568)(11,0.544)};
\addplot[mark=*, mark size=1.8pt, c2, thick, mark options={solid}, only marks] coordinates {(11,0.545)};
\addplot[mark=*, mark size=1pt, c4, mark options={solid}, dotted] coordinates {(12,0.541)(12,0.546)(12,0.557)(12,0.564)(12,0.522)};
\addplot[mark=*, mark size=1.8pt, c4, thick, mark options={solid}, only marks] coordinates {(12,0.546)};
\addplot[mark=*, mark size=1pt, c3, mark options={solid}, dotted] coordinates {(13,0.541)(13,0.552)(13,0.549)(13,0.570)(13,0.523)};
\addplot[mark=*, mark size=1.8pt, c3, thick, mark options={solid}, only marks] coordinates {(13,0.547)};
\addplot[mark=*, mark size=1pt, c3, mark options={solid}, dotted] coordinates {(14,0.544)(14,0.538)(14,0.559)(14,0.554)(14,0.545)};
\addplot[mark=*, mark size=1.8pt, c3, thick, mark options={solid}, only marks] coordinates {(14,0.548)};
\addplot[mark=*, mark size=1pt, c4, mark options={solid}, dotted] coordinates {(15,0.544)(15,0.551)(15,0.562)(15,0.555)(15,0.531)};
\addplot[mark=*, mark size=1.8pt, c4, thick, mark options={solid}, only marks] coordinates {(15,0.549)};
\addplot[mark=*, mark size=1pt, c3, mark options={solid}, dotted] coordinates {(16,0.547)(16,0.547)(16,0.568)(16,0.552)(16,0.536)};
\addplot[mark=*, mark size=1.8pt, c3, thick, mark options={solid}, only marks] coordinates {(16,0.550)};
\addplot[mark=*, mark size=1pt, c3, mark options={solid}, dotted] coordinates {(17,0.558)(17,0.557)(17,0.557)(17,0.557)(17,0.528)};
\addplot[mark=*, mark size=1.8pt, c3, thick, mark options={solid}, only marks] coordinates {(17,0.551)};
\addplot[mark=*, mark size=1pt, c4, mark options={solid}, dotted] coordinates {(18,0.546)(18,0.551)(18,0.557)(18,0.559)(18,0.545)};
\addplot[mark=*, mark size=1.8pt, c4, thick, mark options={solid}, only marks] coordinates {(18,0.552)};
\addplot[mark=*, mark size=1pt, c3, mark options={solid}, dotted] coordinates {(19,0.554)(19,0.559)(19,0.561)(19,0.573)(19,0.528)};
\addplot[mark=*, mark size=1.8pt, c3, thick, mark options={solid}, only marks] coordinates {(19,0.555)};
\addplot[mark=*, mark size=1pt, c4, mark options={solid}, dotted] coordinates {(20,0.544)(20,0.554)(20,0.565)(20,0.568)(20,0.544)};
\addplot[mark=*, mark size=1.8pt, c4, thick, mark options={solid}, only marks] coordinates {(20,0.555)};
\addplot[mark=*, mark size=1pt, c3, mark options={solid}, dotted] coordinates {(21,0.551)(21,0.566)(21,0.562)(21,0.567)(21,0.541)};
\addplot[mark=*, mark size=1.8pt, c3, thick, mark options={solid}, only marks] coordinates {(21,0.557)};
\addplot[mark=*, mark size=1pt, c3, mark options={solid}, dotted] coordinates {(22,0.552)(22,0.558)(22,0.556)(22,0.573)(22,0.556)};
\addplot[mark=*, mark size=1.8pt, c3, thick, mark options={solid}, only marks] coordinates {(22,0.559)};
\addplot[mark=*, mark size=1pt, black, mark options={solid}, dotted] coordinates {(24,0.585)(24,0.580)(24,0.578)(24,0.595)(24,0.566)};
\addplot[mark=*, mark size=1.8pt, black, thick, mark options={solid}, only marks] coordinates {(24,0.581)};
\addplot[mark=*, mark size=1pt, black, mark options={solid}, dotted] coordinates {(26,0.580)(26,0.586)(26,0.583)(26,0.586)(26,0.566)};
\addplot[mark=*, mark size=1.8pt, black, thick, mark options={solid}, only marks] coordinates {(26,0.580)};
\end{axis}
\end{tikzpicture}%
\vskip3pt
\begin{tikzpicture}
\begin{axis}[bplot, ytick distance={0.05}, xticklabels={\color{c1}2\vphantom{0},\color{c1}1\vphantom{0},\color{c1}0\vphantom{0},\color{c2}8\vphantom{0},\color{c3}12\vphantom{0},\color{c2}7\vphantom{0},\color{c3}10\vphantom{0},\color{c3}14\vphantom{0},\color{c3}16\vphantom{0},\color{c3}11\vphantom{0},\color{c2}5\vphantom{0},\color{c3}15\vphantom{0},\color{c2}4\vphantom{0},\color{c4}20\vphantom{0},\color{c2}6\vphantom{0},\color{c3}13\vphantom{0},\color{c3}9\vphantom{0},\color{c2}3\vphantom{0},\color{c4}22\vphantom{0},\color{c4}21\vphantom{0},\color{c4}19\vphantom{0},\color{c3}17\vphantom{0},\color{c3}18\vphantom{0},\color{black}PET\vphantom{0},\color{black}iPET\vphantom{0}}, title={\scriptsize\sffamily Yahoo Questions ($n = 10$)}]
\addplot[decentgrey!80!black, dashed] coordinates {(0,0.553608695652174)(26,0.553608695652174)};
\addplot[mark=*, mark size=1pt, c1, mark options={solid}, dotted] coordinates {(0,0.438)(0,0.372)(0,0.453)(0,0.475)(0,0.334)};
\addplot[mark=*, mark size=1.8pt, c1, thick, mark options={solid}, only marks] coordinates {(0,0.415)};
\addplot[mark=*, mark size=1pt, c1, mark options={solid}, dotted] coordinates {(1,0.499)(1,0.467)(1,0.569)(1,0.514)(1,0.333)};
\addplot[mark=*, mark size=1.8pt, c1, thick, mark options={solid}, only marks] coordinates {(1,0.477)};
\addplot[mark=*, mark size=1pt, c1, mark options={solid}, dotted] coordinates {(2,0.522)(2,0.478)(2,0.552)(2,0.519)(2,0.460)};
\addplot[mark=*, mark size=1.8pt, c1, thick, mark options={solid}, only marks] coordinates {(2,0.506)};
\addplot[mark=*, mark size=1pt, c2, mark options={solid}, dotted] coordinates {(3,0.496)(3,0.496)(3,0.578)(3,0.568)(3,0.418)};
\addplot[mark=*, mark size=1.8pt, c2, thick, mark options={solid}, only marks] coordinates {(3,0.511)};
\addplot[mark=*, mark size=1pt, c3, mark options={solid}, dotted] coordinates {(4,0.489)(4,0.526)(4,0.558)(4,0.518)(4,0.494)};
\addplot[mark=*, mark size=1.8pt, c3, thick, mark options={solid}, only marks] coordinates {(4,0.517)};
\addplot[mark=*, mark size=1pt, c2, mark options={solid}, dotted] coordinates {(5,0.568)(5,0.466)(5,0.548)(5,0.556)(5,0.497)};
\addplot[mark=*, mark size=1.8pt, c2, thick, mark options={solid}, only marks] coordinates {(5,0.527)};
\addplot[mark=*, mark size=1pt, c3, mark options={solid}, dotted] coordinates {(6,0.517)(6,0.555)(6,0.566)(6,0.522)(6,0.475)};
\addplot[mark=*, mark size=1.8pt, c3, thick, mark options={solid}, only marks] coordinates {(6,0.527)};
\addplot[mark=*, mark size=1pt, c3, mark options={solid}, dotted] coordinates {(7,0.504)(7,0.535)(7,0.570)(7,0.580)(7,0.452)};
\addplot[mark=*, mark size=1.8pt, c3, thick, mark options={solid}, only marks] coordinates {(7,0.528)};
\addplot[mark=*, mark size=1pt, c3, mark options={solid}, dotted] coordinates {(8,0.527)(8,0.535)(8,0.598)(8,0.556)(8,0.504)};
\addplot[mark=*, mark size=1.8pt, c3, thick, mark options={solid}, only marks] coordinates {(8,0.544)};
\addplot[mark=*, mark size=1pt, c3, mark options={solid}, dotted] coordinates {(9,0.569)(9,0.573)(9,0.593)(9,0.585)(9,0.487)};
\addplot[mark=*, mark size=1.8pt, c3, thick, mark options={solid}, only marks] coordinates {(9,0.562)};
\addplot[mark=*, mark size=1pt, c2, mark options={solid}, dotted] coordinates {(10,0.583)(10,0.539)(10,0.580)(10,0.587)(10,0.521)};
\addplot[mark=*, mark size=1.8pt, c2, thick, mark options={solid}, only marks] coordinates {(10,0.562)};
\addplot[mark=*, mark size=1pt, c3, mark options={solid}, dotted] coordinates {(11,0.572)(11,0.562)(11,0.608)(11,0.574)(11,0.550)};
\addplot[mark=*, mark size=1.8pt, c3, thick, mark options={solid}, only marks] coordinates {(11,0.573)};
\addplot[mark=*, mark size=1pt, c2, mark options={solid}, dotted] coordinates {(12,0.576)(12,0.597)(12,0.621)(12,0.585)(12,0.506)};
\addplot[mark=*, mark size=1.8pt, c2, thick, mark options={solid}, only marks] coordinates {(12,0.577)};
\addplot[mark=*, mark size=1pt, c4, mark options={solid}, dotted] coordinates {(13,0.545)(13,0.592)(13,0.595)(13,0.578)(13,0.594)};
\addplot[mark=*, mark size=1.8pt, c4, thick, mark options={solid}, only marks] coordinates {(13,0.581)};
\addplot[mark=*, mark size=1pt, c2, mark options={solid}, dotted] coordinates {(14,0.611)(14,0.565)(14,0.597)(14,0.587)(14,0.563)};
\addplot[mark=*, mark size=1.8pt, c2, thick, mark options={solid}, only marks] coordinates {(14,0.585)};
\addplot[mark=*, mark size=1pt, c3, mark options={solid}, dotted] coordinates {(15,0.582)(15,0.591)(15,0.624)(15,0.588)(15,0.538)};
\addplot[mark=*, mark size=1.8pt, c3, thick, mark options={solid}, only marks] coordinates {(15,0.585)};
\addplot[mark=*, mark size=1pt, c3, mark options={solid}, dotted] coordinates {(16,0.606)(16,0.588)(16,0.597)(16,0.588)(16,0.551)};
\addplot[mark=*, mark size=1.8pt, c3, thick, mark options={solid}, only marks] coordinates {(16,0.586)};
\addplot[mark=*, mark size=1pt, c2, mark options={solid}, dotted] coordinates {(17,0.606)(17,0.570)(17,0.604)(17,0.587)(17,0.569)};
\addplot[mark=*, mark size=1.8pt, c2, thick, mark options={solid}, only marks] coordinates {(17,0.587)};
\addplot[mark=*, mark size=1pt, c4, mark options={solid}, dotted] coordinates {(18,0.601)(18,0.577)(18,0.605)(18,0.593)(18,0.571)};
\addplot[mark=*, mark size=1.8pt, c4, thick, mark options={solid}, only marks] coordinates {(18,0.589)};
\addplot[mark=*, mark size=1pt, c4, mark options={solid}, dotted] coordinates {(19,0.584)(19,0.589)(19,0.602)(19,0.569)(19,0.610)};
\addplot[mark=*, mark size=1.8pt, c4, thick, mark options={solid}, only marks] coordinates {(19,0.591)};
\addplot[mark=*, mark size=1pt, c4, mark options={solid}, dotted] coordinates {(20,0.588)(20,0.593)(20,0.610)(20,0.578)(20,0.601)};
\addplot[mark=*, mark size=1.8pt, c4, thick, mark options={solid}, only marks] coordinates {(20,0.594)};
\addplot[mark=*, mark size=1pt, c3, mark options={solid}, dotted] coordinates {(21,0.599)(21,0.582)(21,0.645)(21,0.576)(21,0.580)};
\addplot[mark=*, mark size=1.8pt, c3, thick, mark options={solid}, only marks] coordinates {(21,0.596)};
\addplot[mark=*, mark size=1pt, c3, mark options={solid}, dotted] coordinates {(22,0.615)(22,0.603)(22,0.645)(22,0.606)(22,0.600)};
\addplot[mark=*, mark size=1.8pt, c3, thick, mark options={solid}, only marks] coordinates {(22,0.614)};
\addplot[mark=*, mark size=1pt, black, mark options={solid}, dotted] coordinates {(24,0.636)(24,0.597)(24,0.652)(24,0.634)(24,0.627)};
\addplot[mark=*, mark size=1.8pt, black, thick, mark options={solid}, only marks] coordinates {(24,0.629)};
\addplot[mark=*, mark size=1pt, black, mark options={solid}, dotted] coordinates {(26,0.670)(26,0.663)(26,0.676)(26,0.670)(26,0.679)};
\addplot[mark=*, mark size=1.8pt, black, thick, mark options={solid}, only marks] coordinates {(26,0.672)};
\end{axis}
\end{tikzpicture}%
$\;$%
\begin{tikzpicture}
\begin{axis}[bplot, ytick distance={0.02}, xticklabels={\color{c1}2\vphantom{0},\color{c1}0\vphantom{0},\color{c1}1\vphantom{0},\color{c3}10\vphantom{0},\color{c3}14\vphantom{0},\color{c3}12\vphantom{0},\color{c2}8\vphantom{0},\color{c2}7\vphantom{0},\color{c3}11\vphantom{0},\color{c3}16\vphantom{0},\color{c4}21\vphantom{0},\color{c2}3\vphantom{0},\color{c2}4\vphantom{0},\color{c4}19\vphantom{0},\color{c4}20\vphantom{0},\color{c4}22\vphantom{0},\color{c3}15\vphantom{0},\color{c3}9\vphantom{0},\color{c2}5\vphantom{0},\color{c2}6\vphantom{0},\color{c3}13\vphantom{0},\color{c3}17\vphantom{0},\color{c3}18\vphantom{0},\color{black}PET\vphantom{0},\color{black}iPET\vphantom{0}}, title={\scriptsize\sffamily Yahoo Questions ($n = 100$)}]
\addplot[decentgrey!80!black, dashed] coordinates {(0,0.6443278260869564)(26,0.6443278260869564)};
\addplot[mark=*, mark size=1pt, c1, mark options={solid}, dotted] coordinates {(0,0.611)(0,0.601)(0,0.624)(0,0.602)(0,0.601)};
\addplot[mark=*, mark size=1.8pt, c1, thick, mark options={solid}, only marks] coordinates {(0,0.608)};
\addplot[mark=*, mark size=1pt, c1, mark options={solid}, dotted] coordinates {(1,0.631)(1,0.619)(1,0.618)(1,0.632)(1,0.616)};
\addplot[mark=*, mark size=1.8pt, c1, thick, mark options={solid}, only marks] coordinates {(1,0.623)};
\addplot[mark=*, mark size=1pt, c1, mark options={solid}, dotted] coordinates {(2,0.632)(2,0.618)(2,0.635)(2,0.622)(2,0.620)};
\addplot[mark=*, mark size=1.8pt, c1, thick, mark options={solid}, only marks] coordinates {(2,0.625)};
\addplot[mark=*, mark size=1pt, c3, mark options={solid}, dotted] coordinates {(3,0.627)(3,0.615)(3,0.641)(3,0.640)(3,0.629)};
\addplot[mark=*, mark size=1.8pt, c3, thick, mark options={solid}, only marks] coordinates {(3,0.630)};
\addplot[mark=*, mark size=1pt, c3, mark options={solid}, dotted] coordinates {(4,0.640)(4,0.629)(4,0.636)(4,0.621)(4,0.643)};
\addplot[mark=*, mark size=1.8pt, c3, thick, mark options={solid}, only marks] coordinates {(4,0.634)};
\addplot[mark=*, mark size=1pt, c3, mark options={solid}, dotted] coordinates {(5,0.635)(5,0.636)(5,0.654)(5,0.639)(5,0.614)};
\addplot[mark=*, mark size=1.8pt, c3, thick, mark options={solid}, only marks] coordinates {(5,0.636)};
\addplot[mark=*, mark size=1pt, c2, mark options={solid}, dotted] coordinates {(6,0.643)(6,0.625)(6,0.645)(6,0.630)(6,0.640)};
\addplot[mark=*, mark size=1.8pt, c2, thick, mark options={solid}, only marks] coordinates {(6,0.637)};
\addplot[mark=*, mark size=1pt, c2, mark options={solid}, dotted] coordinates {(7,0.643)(7,0.638)(7,0.643)(7,0.629)(7,0.647)};
\addplot[mark=*, mark size=1.8pt, c2, thick, mark options={solid}, only marks] coordinates {(7,0.640)};
\addplot[mark=*, mark size=1pt, c3, mark options={solid}, dotted] coordinates {(8,0.639)(8,0.622)(8,0.661)(8,0.644)(8,0.642)};
\addplot[mark=*, mark size=1.8pt, c3, thick, mark options={solid}, only marks] coordinates {(8,0.641)};
\addplot[mark=*, mark size=1pt, c3, mark options={solid}, dotted] coordinates {(9,0.634)(9,0.637)(9,0.655)(9,0.649)(9,0.636)};
\addplot[mark=*, mark size=1.8pt, c3, thick, mark options={solid}, only marks] coordinates {(9,0.642)};
\addplot[mark=*, mark size=1pt, c4, mark options={solid}, dotted] coordinates {(10,0.659)(10,0.642)(10,0.651)(10,0.640)(10,0.648)};
\addplot[mark=*, mark size=1.8pt, c4, thick, mark options={solid}, only marks] coordinates {(10,0.648)};
\addplot[mark=*, mark size=1pt, c2, mark options={solid}, dotted] coordinates {(11,0.660)(11,0.641)(11,0.657)(11,0.640)(11,0.648)};
\addplot[mark=*, mark size=1.8pt, c2, thick, mark options={solid}, only marks] coordinates {(11,0.649)};
\addplot[mark=*, mark size=1pt, c2, mark options={solid}, dotted] coordinates {(12,0.649)(12,0.644)(12,0.658)(12,0.646)(12,0.650)};
\addplot[mark=*, mark size=1.8pt, c2, thick, mark options={solid}, only marks] coordinates {(12,0.650)};
\addplot[mark=*, mark size=1pt, c4, mark options={solid}, dotted] coordinates {(13,0.651)(13,0.644)(13,0.660)(13,0.653)(13,0.641)};
\addplot[mark=*, mark size=1.8pt, c4, thick, mark options={solid}, only marks] coordinates {(13,0.650)};
\addplot[mark=*, mark size=1pt, c4, mark options={solid}, dotted] coordinates {(14,0.657)(14,0.651)(14,0.661)(14,0.648)(14,0.641)};
\addplot[mark=*, mark size=1.8pt, c4, thick, mark options={solid}, only marks] coordinates {(14,0.652)};
\addplot[mark=*, mark size=1pt, c4, mark options={solid}, dotted] coordinates {(15,0.644)(15,0.659)(15,0.661)(15,0.643)(15,0.653)};
\addplot[mark=*, mark size=1.8pt, c4, thick, mark options={solid}, only marks] coordinates {(15,0.652)};
\addplot[mark=*, mark size=1pt, c3, mark options={solid}, dotted] coordinates {(16,0.633)(16,0.650)(16,0.681)(16,0.652)(16,0.653)};
\addplot[mark=*, mark size=1.8pt, c3, thick, mark options={solid}, only marks] coordinates {(16,0.654)};
\addplot[mark=*, mark size=1pt, c3, mark options={solid}, dotted] coordinates {(17,0.655)(17,0.641)(17,0.671)(17,0.651)(17,0.655)};
\addplot[mark=*, mark size=1.8pt, c3, thick, mark options={solid}, only marks] coordinates {(17,0.655)};
\addplot[mark=*, mark size=1pt, c2, mark options={solid}, dotted] coordinates {(18,0.664)(18,0.653)(18,0.660)(18,0.647)(18,0.653)};
\addplot[mark=*, mark size=1.8pt, c2, thick, mark options={solid}, only marks] coordinates {(18,0.655)};
\addplot[mark=*, mark size=1pt, c2, mark options={solid}, dotted] coordinates {(19,0.661)(19,0.654)(19,0.656)(19,0.652)(19,0.657)};
\addplot[mark=*, mark size=1.8pt, c2, thick, mark options={solid}, only marks] coordinates {(19,0.656)};
\addplot[mark=*, mark size=1pt, c3, mark options={solid}, dotted] coordinates {(20,0.667)(20,0.652)(20,0.673)(20,0.645)(20,0.654)};
\addplot[mark=*, mark size=1.8pt, c3, thick, mark options={solid}, only marks] coordinates {(20,0.658)};
\addplot[mark=*, mark size=1pt, c3, mark options={solid}, dotted] coordinates {(21,0.667)(21,0.655)(21,0.667)(21,0.652)(21,0.656)};
\addplot[mark=*, mark size=1.8pt, c3, thick, mark options={solid}, only marks] coordinates {(21,0.659)};
\addplot[mark=*, mark size=1pt, c3, mark options={solid}, dotted] coordinates {(22,0.667)(22,0.663)(22,0.673)(22,0.645)(22,0.673)};
\addplot[mark=*, mark size=1.8pt, c3, thick, mark options={solid}, only marks] coordinates {(22,0.664)};
\addplot[mark=*, mark size=1pt, black, mark options={solid}, dotted] coordinates {(24,0.682)(24,0.680)(24,0.686)(24,0.673)(24,0.680)};
\addplot[mark=*, mark size=1.8pt, black, thick, mark options={solid}, only marks] coordinates {(24,0.680)};
\addplot[mark=*, mark size=1pt, black, mark options={solid}, dotted] coordinates {(26,0.675)(26,0.679)(26,0.681)(26,0.673)(26,0.679)};
\addplot[mark=*, mark size=1.8pt, black, thick, mark options={solid}, only marks] coordinates {(26,0.677)};
\end{axis}
\end{tikzpicture}%
\caption{Performance of individual patterns, \pet{} and \ipet{} on all tasks considered. Accuracy is shown on the $y$-axis; the $x$-axis shows individual pattern ids where color is used to distinguish the different pattern categories (\textcolor{c1}{\textsc{Null}}, \textcolor{c2}{\textsc{Punc}}, \textcolor{c3}{\textsc{Prompt}}, \textcolor{c4}{\textsc{Q\&A}}). Small bullets (\tikz[baseline=-0.6ex]{\fill[black, inner sep=0]  circle(1.4pt);}) correspond to individual training sets, large bullets (\tikz[baseline=-0.6ex]{\fill[black, thick, inner sep=0]  circle(2.2pt);}) correspond to average performance. Average performance across all patterns is shown as a dashed gray line.}
\label{fig:performance-default-patterns}
\end{figure*}
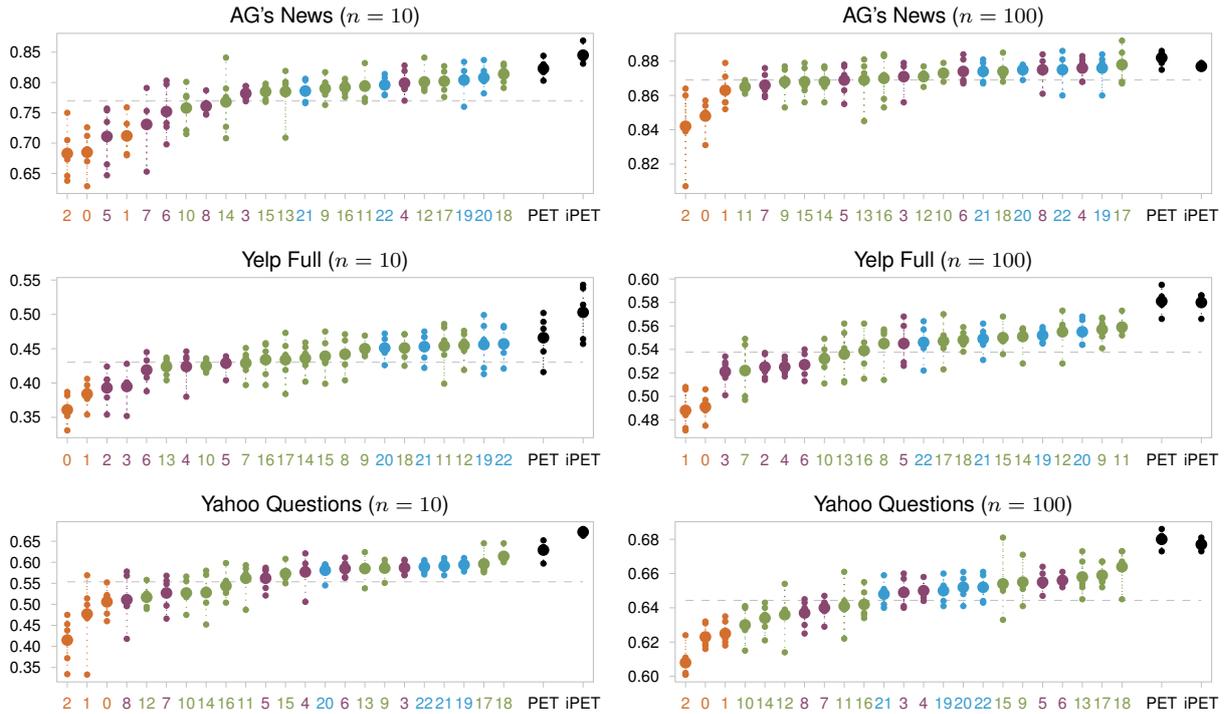

Even slightly different patterns can lead to very different performance \citep[][i.a.]{jiang2019know,schick2020exploiting,schick2021selfdiagnosis,webson2021promptbased,sanh2021multitask} and popular model selection criteria can not reliably identify patterns that achieve similar performance to the best one \citep{perez2021true}. We thus investigate to what extent \pet{} can eliminate the need to find the best instruction even in extreme settings where there are dozens of candidates to choose from.

\paragraph{Setup} Using our default setup, we train individual models for each PVP and a final \pet{} model; we also train an \ipet{} model for 3 iterations.

\paragraph{Results} Performance of individual models for each pattern and of the distilled models obtained using \pet{} and \ipet{} is shown in Figure~\ref{fig:performance-default-patterns}. Interestingly, sorting all pattern groups by their average performance gives the exact same order for each task and training set size: \textsc{Null} patterns clearly perform worst, followed by \textsc{Punc} and \textsc{Prompt}; \textsc{Q\&A} gives the best average results. Contrary to findings of \citet{logan2021cutting}, this shows that LMs can benefit a lot from manually written instructions even if combined with finetuning. 

Crucially, \pet{}'s performance is much higher than average performance of individual patterns; further, it consistently outperforms even the best pattern, verifying that \pet{} indeed removes the need to find \emph{the} best pattern. While \ipet{} gives clear improvements for $n = 10$, it performs worse than \pet{} for $n = 100$. The reason for this may be that we use a much smaller set of unlabeled examples than prior work \citep{schick2020exploiting,schick2020just}.

\question{Does performance of different patterns transfer across models?}
\label{q2}

While our results for Q\ref{q1} show a consistent order of pattern groups for different training set sizes and tasks, an important question for real-world applications is whether the same finding also holds for different model sizes, and even entirely different models.

\paragraph{Setup} We consider BERT \citep{devlin2018bert} RoBERTa \citep{liu2019roberta} and ALBERT \citep{lan2019albert} as underlying language models;\footnote{For Yahoo, we do not consider BERT as it uses a vocabulary that does not assign a single token to each verbalization.} we experiment with the base and large variants. For each model and size, we repeat the exact same experiment as for Q\ref{q1}.

\paragraph{Results} Figure~\ref{fig:performance-models} shows the performance of each pattern group (i.e., average performance of all individual patterns in this group) and \pet{};  scores are normalized so that the best-performing approach for each task, training set size and model gets a score of $1.0$. Except for very few exceptions, our findings from Q\ref{q1} regarding the relative performance of pattern groups and \pet{} (\textsc{Null} < \textsc{Punc} < \textsc{Prompt} < \textsc{Q\&A} < \pet{}) also hold for different models and sizes. In general, the performance of individual patterns strongly correlates between different models and sizes (Spearman's $\rho \geq 0.7$ except in one case).

\pgfplotsset{
	heatmap/.style={
		width=0.36\linewidth,
		height=0.13\textheight,
		view={0}{90},
		colormap = {blackwhite}{color(0cm)  = (white);color(0.5cm) = (white!90!c0); color(1cm) = (c0)},
		point meta min=0.52,
		point meta max=1,
		ymin=-0.5,
		ymax=5.5,
		xmin=-0.5,
		xmax=4.5,
		scale only axis,
		clip=false,
		xtick={0,1,2,3,4},
		xticklabels={\textsc{Null}, \textsc{Punc}, \textsc{Prompt}, \textsc{Q\&A}, \pet{}},
		yticklabels={RoBERTa (base), RoBERTa (large), BERT (base), BERT (large), ALBERT (base), ALBERT (large)},
		ytick={0,1,2,3,4,5},
		xtick pos=left,
		ytick pos=left,
		ylabel near ticks,
		xlabel near ticks,
		x tick label style={rotate=45,anchor=east},
		tick align=outside,
		major tick length=0.075cm,
		tick label style={font=\sffamily\tiny},
		every node near coord/.append style={anchor=center,color=white,font=\sansmath\sffamily\tiny, /pgf/number format/.cd, fixed, fixed zerofill, precision=2, /tikz/.cd},
	},
	heatmap-agnews/.style={
		heatmap,			
	},
	heatmap-yelp/.style={
		heatmap,
	},
	heatmap-yahoo/.style={
		heatmap, ymax=3.5, height=0.1\textheight
	},
}

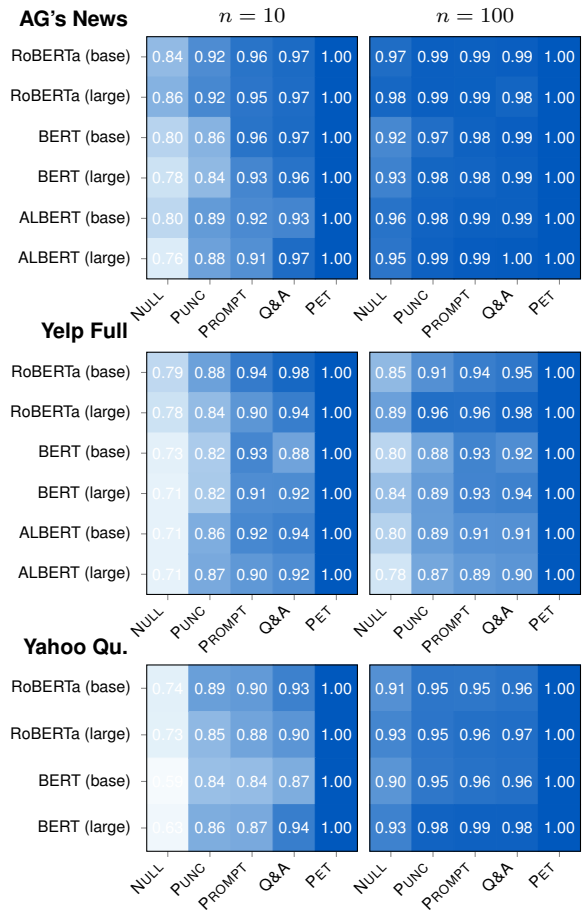
\begin{figure}
	\centering
	\begin{tikzpicture}
	\begin{axis}[heatmap-agnews]
	\addplot[matrix plot,point meta=explicit, nodes near coords] table [x=pattern_id, y=model_id, meta=value, col sep=comma] {heatmap_agnews_10_grouped.csv};
	\node[anchor=south]() at (2,-0.6) {\scriptsize $n = 10$};
	\node[anchor=south east]() at (-0.7,-0.5) {\scriptsize\sffamily\textbf{AG's News}};
	\end{axis}
	\end{tikzpicture}%
	~\hspace{-0.4cm}%
	\begin{tikzpicture}
	\begin{axis}[heatmap-agnews, ymajorticks=false]
	\addplot[matrix plot,point meta=explicit, nodes near coords] table [x=pattern_id, y=model_id, meta=value, col sep=comma] {heatmap_agnews_100_grouped.csv};
	\node[anchor=south]() at (2,-0.6) {\scriptsize $n = 100$};	
	\end{axis}
	\end{tikzpicture}
	\vskip-16pt
	\begin{tikzpicture}
	\begin{axis}[heatmap-yelp]
	\addplot[matrix plot,point meta=explicit, nodes near coords] table [x=pattern_id, y=model_id, meta=value, col sep=comma] {heatmap_yelp-full_10_grouped.csv};
	\node[anchor=south east]() at (-0.7,-0.5) {\scriptsize\sffamily\textbf{Yelp Full}};
	\end{axis}
	\end{tikzpicture}%
	~\hspace{-0.4cm}%
	\begin{tikzpicture}
	\begin{axis}[heatmap-yelp, ymajorticks=false]
	\addplot[matrix plot,point meta=explicit, nodes near coords] table [x=pattern_id, y=model_id, meta=value, col sep=comma] {heatmap_yelp-full_100_grouped.csv};
	\end{axis}
	\end{tikzpicture}
	\vskip-16pt
	\begin{tikzpicture}
	\begin{axis}[heatmap-yahoo]
	\addplot[matrix plot,point meta=explicit, nodes near coords] table [x=pattern_id, y=model_id, meta=value, col sep=comma] {heatmap_yahoo_10_grouped.csv};
	\node[anchor=south east]() at (-0.7,-0.5) {\scriptsize\sffamily\textbf{Yahoo Qu.}};
	\end{axis}
	\end{tikzpicture}%
	~\hspace{-0.4cm}%
	\begin{tikzpicture}
	\begin{axis}[heatmap-yahoo, ymajorticks=false]
	\addplot[matrix plot,point meta=explicit, nodes near coords] table [x=pattern_id, y=model_id, meta=value, col sep=comma] {heatmap_yahoo_100_grouped.csv};
	\end{axis}
	\end{tikzpicture}
	\caption{Relative performance of individual pattern groups and \pet{} for different models and sizes. Scores are normalized so that the best performance for each task, number of examples, model and size is $1.0$.}
	\label{fig:performance-models}
\end{figure}

\question{Does \pet{} still work if some PVPs are not well understood?}
\label{q3}

While Q\ref{q1} and Q\ref{q2} have shown that \pet{} performs even better than the best PVP if a large set of high-quality PVPs is given, a potential concern is that performance may be much worse if the LM fails to understand a fair amount of patterns and verbalizers (e.g., because they are in a very different style from the model's pretraining data). For real-world scenarios, it is thus relevant to know how such ``bad'' PVPs affect the performance of \pet{}.

\begin{figure}
	\centering
	\begin{tikzpicture}
	\begin{axis}[bplot,
	width=1.08\linewidth, 
	legend style={draw=darkgrey, fill=white!75, text opacity =1, fill opacity=0.8, at={(0.025,0.05)},anchor=south west, font=\sffamily\scriptsize},
	legend cell align=left,
	legend columns=3,
	xtick={0, 2, 4, 6, 8, 10, 12, 14, 16, 18, 20, 22, 24},
	ymin=0, ymax=1,
	xticklabels={0, 2, 4, 6, 8, 10, 12, 14, 16, 18, 20, \textsc{NP}+P, \textsc{NP}},
	grid=major, clip=false,
	major grid style={line width=.2pt,draw=decentgrey},
	minor tick style={decentgrey!0},
	major tick style={decentgrey}, ytick distance={0.1}, height=0.2\textheight, enlarge y limits=0, enlarge x limits=0.05]
	\addplot[mark=*,  mark options={solid}, mark size=1.5pt, line width=0.6pt,solid,color=c0] table[x=key,y=agnews,col sep=comma]{random_10.csv};
	\addlegendentry{AG's}
	\addplot[mark=*, forget plot, mark options={solid}, mark size=1.5pt, line width=0.6pt,solid,color=c0, dotted] table[x=key,y=agnews,col sep=comma]{random_10_special.csv};
	
	\addplot[mark=triangle*, mark options={solid}, mark size=1.5pt, line width=0.6pt,solid,color=c1] table[x=key,y=yelp-full,col sep=comma]{random_10.csv};
	\addlegendentry{Yelp}
	\addplot[mark=triangle*, forget plot,  mark options={solid}, mark size=1.5pt, line width=0.6pt,solid,color=c1, dotted] table[x=key,y=yelp-full,col sep=comma]{random_10_special.csv};
	
	\addplot[mark=pentagon*,  mark options={solid}, mark size=1.5pt, line width=0.6pt,solid,color=c2] table[x=key,y=yahoo,col sep=comma]{random_10.csv};
	\addlegendentry{Yahoo}
	\addplot[mark=pentagon*, forget plot, mark options={solid}, mark size=1.5pt, line width=0.6pt,solid,color=c2, dotted] table[x=key,y=yahoo,col sep=comma]{random_10_special.csv};
	
	\node[anchor=south]() at (12,1) {\scriptsize $n = 10$};
	\end{axis}
	\end{tikzpicture}%
	
	\begin{tikzpicture}
	\begin{axis}[bplot,
	width=1.08\linewidth, 
	legend style={draw=darkgrey, fill=white!75, text opacity =1, fill opacity=0.8, at={(0.025,0.05)},anchor=south west, font=\sffamily\scriptsize},
	legend cell align=left,
	legend columns=3,
	xtick={0, 2, 4, 6, 8, 10, 12, 14, 16, 18, 20, 22, 24},
	ymin=0, ymax=1,
	xticklabels={0, 2, 4, 6, 8, 10, 12, 14, 16, 18, 20, \textsc{NP}+P, \textsc{NP}},
	grid=major,
	major grid style={line width=.2pt,draw=decentgrey},
	minor tick style={decentgrey!0},
	clip=false,
	major tick style={decentgrey}, ytick distance={0.1}, height=0.2\textheight, enlarge y limits=0, enlarge x limits=0.05]
	\addplot[mark=*,  mark options={solid}, mark size=1.5pt, line width=0.6pt,solid,color=c0] table[x=key,y=agnews,col sep=comma]{random_100.csv};
	\addlegendentry{AG's}
	\addplot[mark=*, forget plot, mark options={solid}, mark size=1.5pt, line width=0.6pt,solid,color=c0, dotted] table[x=key,y=agnews,col sep=comma]{random_100_special.csv};
	
	\addplot[mark=triangle*,  mark options={solid}, mark size=1.5pt, line width=0.6pt,solid,color=c1] table[x=key,y=yelp-full,col sep=comma]{random_100.csv};
	\addlegendentry{Yelp}
	\addplot[mark=triangle*, forget plot, mark options={solid}, mark size=1.5pt, line width=0.6pt,solid,color=c1, dotted] table[x=key,y=yelp-full,col sep=comma]{random_100_special.csv};
	
	\addplot[mark=pentagon*,  mark options={solid}, mark size=1.5pt, line width=0.6pt,solid,color=c2] table[x=key,y=yahoo,col sep=comma]{random_100.csv};
	\addlegendentry{Yahoo}
	\addplot[mark=pentagon*, forget plot, mark options={solid}, mark size=1.5pt, line width=0.6pt,solid,color=c2, dotted] table[x=key,y=yahoo,col sep=comma]{random_100_special.csv};
	
	\node[anchor=south]() at (12,1) {\scriptsize $n = 100$};
	\end{axis}
	\end{tikzpicture}
	\caption{Performance of \pet{} with three randomly selected patterns when adding noise PVPs; the $x$-axis shows the number of noise PVPs added. We also show performance of using \emph{only} noise PVPs with \pet{} (\textsc{NP+P)} and their average performance (NP).}
	\label{fig:performance-random-patterns}
\end{figure}
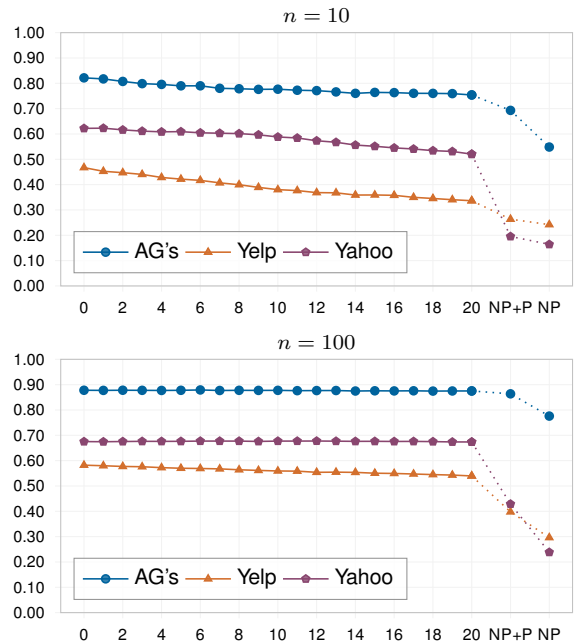

\paragraph{Setup} It is difficult to obtain large quantities of bad instructions as they might occur in real-world scenarios. As a proxy, we resort to \emph{noise patterns} that add random tokens to the input, serving as a lower bound for truly bad patterns. In concrete terms, we add up to three randomly sampled tokens before and after the input.\footnote{If there are multiple input texts, we shuffle their order and additionally add 0--3 tokens in between them.} We also create \emph{noise verbalizers} by assigning uniformly selected tokens to each output class. Using this process, we obtain 20 different intentionally bad PVPs per task. For each task, we start with 3 randomly selected, high-quality patterns from our original set of manually designed instructions, add noise PVPs one by one and investigate the effect on performance.

\paragraph{Results} The effect of adding noise PVPs is shown in Figure~\ref{fig:performance-random-patterns}. Interestingly, for both $n = 10$ and $n = 100$, performance remains almost constant even if more than half of the used PVPs are noise PVPs, demonstrating that \pet{} is very robust to ``bad'' instructions. Figure~\ref{fig:performance-random-patterns} also shows performance when using \emph{only} noise PVPs; except for AG's News with $n = 100$, this leads to substantially worse performance than \pet{} with a small amount of manually designed instructions.

\question{How many patterns are required for good performance?}
\label{q4}

Orthogonal to Q\ref{q3}, it is also important to know how many \emph{high-quality} prompts are at least required to achieve satisfactory performance. This is of great practical importance because coming up with dozens of different PVPs for a task may take a significant amount of time, that could otherwise be spent annotating further examples.

\paragraph{Setup} We generate 10 random permutations of all 23 patterns per task. For each permutation and training set, we use the same setup as in Q\ref{q1} to compute the average performance obtained with \pet{} when using only the first $i$ patterns, where $i$ ranges from $1$ to $5$.

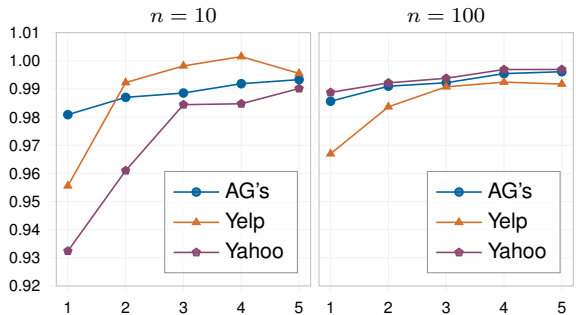
\begin{figure}
		\centering
		\begin{tikzpicture}
		\begin{axis}[bplot,
		width=0.64\linewidth, 
		legend style={draw=darkgrey, fill=white!75, text opacity =1, fill opacity=0.8, at={(0.95,0.05)},anchor=south east, font=\sffamily\scriptsize},
		legend cell align=left,
		legend columns=1,
		xtick={1,2,3,4,5},
		grid=major, clip=false,
		major grid style={line width=.2pt,draw=decentgrey},
		minor tick style={decentgrey!0},
		x tick label style={font=\sansmath\sffamily\tiny, inner xsep=0},
		ymin=0.92, ymax=1.01,
		major tick style={decentgrey}, ytick distance={0.01}, height=0.2\textheight, enlarge y limits=0, enlarge x limits=0.05]
		\addplot[mark=*,  mark options={solid}, mark size=1.5pt, line width=0.6pt,solid,color=c0] coordinates {
			(1, 0.9808552513866524)
			(2, 0.9870101986044012)
			(3, 0.9885489354088388)
			(4, 0.9918590087672213)
			(5, 0.9933082841295402)
		};
		\addlegendentry{AG's}
		
		\addplot[mark=triangle*, mark options={solid}, mark size=1.5pt, line width=0.6pt,solid,color=c1] coordinates {
			(1, 0.9555467003165747)
			(2, 0.9922516658918333)
			(3, 0.9981846760089439)
			(4, 1.0014832525292776)
			(5, 0.9955059662172632)	
		};
		\addlegendentry{Yelp}
		
		\addplot[mark=pentagon*,  mark options={solid}, mark size=1.5pt, line width=0.6pt,solid,color=c2] coordinates {
			(1, 0.9323967779699813)
			(2, 0.9609826842096066)
			(3, 0.9844140413412312)
			(4, 0.9847285562020585)
			(5, 0.9901452010274152)
		};
		\addlegendentry{Yahoo}
		
		\node[anchor=south]() at (3,1.01) {\scriptsize $n = 10$};
		\end{axis}
		\end{tikzpicture}%
		~%
		\begin{tikzpicture}
		\begin{axis}[bplot,
		width=0.64\linewidth, 
		legend style={draw=darkgrey, fill=white!75, text opacity =1, fill opacity=0.8, at={(0.95,0.05)},anchor=south east, font=\sffamily\scriptsize},
		legend cell align=left,
		legend columns=1,
		xtick={1,2,3,4,5},
		ymin=0.92, ymax=1.01,
		x tick label style={font=\sansmath\sffamily\tiny, inner xsep=0},
		grid=major, clip=false,
		major grid style={line width=.2pt,draw=decentgrey},
		minor tick style={decentgrey!0},
		ymajorticks=false,
		major tick style={decentgrey}, ytick distance={0.01}, height=0.2\textheight, enlarge y limits=0, enlarge x limits=0.05]
		\addplot[mark=*,  mark options={solid}, mark size=1.5pt, line width=0.6pt,solid,color=c0] coordinates {
			(1, 0.9856273361780954)
			(2, 0.9909695345837056)
			(3, 0.9921934438159505)
			(4, 0.9954516886639541)
			(5, 0.9961132612219247)
		};
		\addlegendentry{AG's}
		
		\addplot[mark=triangle*, mark options={solid}, mark size=1.5pt, line width=0.6pt,solid,color=c1] coordinates {
			(1, 0.9669515669515668)
			(2, 0.9836467236467238)
			(3, 0.9907502374169043)
			(4, 0.9924216524216525)
			(5, 0.9917378917378917)
		};
		\addlegendentry{Yelp}
		
		\addplot[mark=pentagon*,  mark options={solid}, mark size=1.5pt, line width=0.6pt,solid,color=c2] coordinates {
			(1, 0.9887699053021044)
			(2, 0.9921438234479165)
			(3, 0.9937900347171288)
			(4, 0.9969031669193033)
			(5, 0.9969194660407806)
		};
		\addlegendentry{Yahoo}
		
		\node[anchor=south]() at (3,1.01) {\scriptsize $n = 100$};
		\end{axis}
		\end{tikzpicture}%
		
	\caption{Relative performance of \pet{} with only a subset of patterns compared to that achieved using all 23 manually designed patterns. The $x$-axis shows the number of patterns used.}
	\label{fig:performance-num-patterns}
\end{figure}

\paragraph{Results} Average performance of \pet{} trained with the first $i$ patterns is shown in Figure~\ref{fig:performance-num-patterns}, relative to the performance of \pet{} trained with all 23 patterns. For all tasks and training set sizes, as little as four different patterns are already sufficient to achieve performance very close to that of \pet{} trained with all 23 patterns. Surprisingly, \pet{}'s performance is much higher than the average performance of a model trained on individual patterns even with $i = 1$. This indicates that the process of knowledge distillation using unlabeled data is also beneficial when using only a single instruction.

\question{How important are values for other hyperparameters?}
\label{q5}

For true few-shot settings, the same set of hyperparameter values should ideally achieve good performance across different tasks; this enables us to adopt these values for new tasks without requiring manual tuning on task-specific validation sets. We therefore investigate how different choices for common hyperparameters affect the performance of \pet{}; in concrete terms, we consider learning rate, training steps and batch size.

\paragraph{Setup} Based on choices found in previous work, we consider different learning rates from the set $\{10^{-4}, {5\cdot 10^{-4}}, 10^{-5}, 5 \cdot 10^{-5}, 10^{-6} \}$, training steps from $\{10, 25, 50, 100, 250, 500, 1000\}$ and batch sizes from $\{1, 2, 4, 8, 16, 32\}$. Learning rate and batch size are changed for the individual models and the final classifier simultaneously; the number of training steps is varied only for individual models. We modify each hyperparameter independently, keeping all other parameters at their default values (i.e., a learning rate of $10^{-5}$, 100 steps and a batch size of 4).

\paragraph{Results} All results are shown in Figure~\ref{fig:performance-hyperparams}. For training steps and batch size, performance is relatively stable across a wide range of different values, with more steps and larger batch sizes typically leading to slightly better performance (especially for $n=100$). Learning rate clearly has the strongest impact on performance, but values of $10^{-5}$ and $5\cdot 10^{-5}$ consistently give the best results across all tasks considered; those are also the values typically used for finetuning in prior work \citep{devlin2018bert,liu2019roberta}.

\pgfplotsset{
	hplot/.style={
		axis line style={decentgrey!80!black},
		major tick style={decentgrey!80!black},
		xtick pos=left,
		ytick pos=left,
		ylabel near ticks,
		xlabel near ticks,
		tick align=outside,
		enlarge x limits=0.08,
		title style={yshift=-1.5ex},
		enlarge y limits=0,
		grid=major, clip=false,
		major grid style={line width=.2pt,draw=decentgrey},
		major tick length=0.075cm,
		width = 0.64\linewidth,
		height = 0.2\textheight,
		log ticks with fixed point,
		x tick label style={font=\sansmath\sffamily\tiny, inner xsep=0},
		y tick label style={font=\sansmath\sffamily\tiny, /pgf/number format/.cd, fixed, fixed zerofill, precision=2, /tikz/.cd},
	},
}

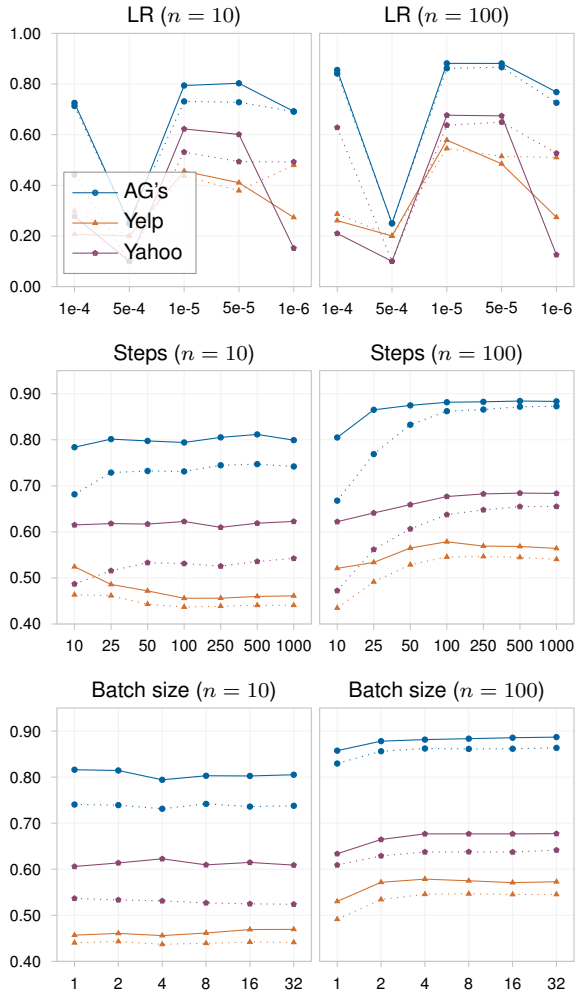
\begin{figure}
	
	\begin{tikzpicture}
	\begin{axis}[hplot, ymin=0, ymax=1,
	legend style={draw=darkgrey, fill=white!75, text opacity =1, fill opacity=0.8, at={(0.025,0.05)},anchor=south west, font=\sffamily\scriptsize},
	legend cell align=left,
	legend columns=1,
	xtick=data, xticklabels={1e-4,5e-4,1e-5,5e-5,1e-6},  title={\scriptsize\sffamily LR ($n = 10$)}]
	\addplot[mark=*, mark size=1pt, c0, mark options={solid}] coordinates {(0, 0.7255263157894738) (1, 0.25) (2, 0.7941315789473684) (3, 0.802921052631579) (4, 0.6919736842105262)};
	\addlegendentry{AG's};
	\addplot[mark=*, forget plot, mark size=1pt, c0, mark options={solid}, dotted] coordinates {(0, 0.713336842105263) (1, 0.25) (2, 0.7313105263157895) (3, 0.7279) (4, 0.6910052631578947)};
	\addplot[mark=triangle*, mark size=1pt, c1, mark options={solid}] coordinates {(0, 0.20696000000000003) (1, 0.2) (2, 0.4558999999999999) (3, 0.41036) (4, 0.27318)};
	\addlegendentry{Yelp};
	\addplot[mark=triangle*, forget plot, mark size=1pt, c1, mark options={solid}, dotted] coordinates {(0, 0.29923200000000005) (1, 0.2) (2, 0.43684399999999995) (3, 0.379004) (4, 0.47940000000000005)};
	\addplot[mark=pentagon*, mark size=1pt, c2, mark options={solid}] coordinates {(0, 0.27790000000000004) (1, 0.1) (2, 0.6225799999999999) (3, 0.601) (4, 0.15174)};
	\addlegendentry{Yahoo};
	\addplot[mark=pentagon*, forget plot, mark size=1pt, c2, mark options={solid}, dotted] coordinates {(0, 0.44066400000000006) (1, 0.1) (2, 0.5312080000000001) (3, 0.493628) (4, 0.49250799999999995)};
	\end{axis}
	\end{tikzpicture}%
~%
\begin{tikzpicture}
	\begin{axis}[hplot, ymin=0, ymax=1, ymajorticks=false, xtick=data, xticklabels={1e-4,5e-4,1e-5,5e-5,1e-6},  title={\scriptsize\sffamily LR ($n = 100$)}]
	\addplot[mark=*, mark size=1pt, c0, mark options={solid}] coordinates {(0, 0.8554210526315791) (1, 0.25) (2, 0.8814736842105262) (3, 0.8813157894736843) (4, 0.768078947368421)};
	\addplot[mark=*, mark size=1pt, c0, mark options={solid}, dotted] coordinates {(0, 0.8411947368421051) (1, 0.25) (2, 0.8620789473684211) (3, 0.8663157894736842) (4, 0.7260105263157894)};
	\addplot[mark=triangle*, mark size=1pt, c1, mark options={solid}] coordinates {(0, 0.26070000000000004) (1, 0.2) (2, 0.5785) (3, 0.48494000000000004) (4, 0.27404)};
	\addplot[mark=triangle*, mark size=1pt, c1, mark options={solid}, dotted] coordinates {(0, 0.287096) (1, 0.2) (2, 0.545572) (3, 0.514552) (4, 0.510056)};
	\addplot[mark=pentagon*, mark size=1pt, c2, mark options={solid}] coordinates {(0, 0.21014) (1, 0.1) (2, 0.67676) (3, 0.6742600000000001) (4, 0.12574000000000002)};
	\addplot[mark=pentagon*, mark size=1pt, c2, mark options={solid}, dotted] coordinates {(0, 0.62828) (1, 0.1) (2, 0.637436) (3, 0.649416) (4, 0.526704)};
	\end{axis}
	\end{tikzpicture}%
	\vskip3pt
	
	\begin{tikzpicture}
	\begin{axis}[hplot, ymin=0.4, ytick distance={0.1}, ymax=0.95, xtick=data, xticklabels={10,25,50,100,250,500,1000},  title={\scriptsize\sffamily Steps ($n = 10$)}]
	\addplot[mark=*, mark size=1pt, c0, mark options={solid}] coordinates {(0, 0.7838157894736841) (1, 0.8014210526315789) (2, 0.7975) (3, 0.7941315789473684) (4, 0.8052105263157895) (5, 0.8116052631578947) (6, 0.7991578947368422)};
	\addplot[mark=*, mark size=1pt, c0, mark options={solid}, dotted] coordinates {(0, 0.6815210526315789) (1, 0.728757894736842) (2, 0.7322684210526316) (3, 0.7313105263157895) (4, 0.7445736842105263) (5, 0.7471526315789474) (6, 0.742)};
	\addplot[mark=triangle*, mark size=1pt, c1, mark options={solid}] coordinates {(0, 0.52436) (1, 0.48586) (2, 0.47187999999999997) (3, 0.4558999999999999) (4, 0.45603999999999995) (5, 0.45990000000000003) (6, 0.46105999999999997)};
	\addplot[mark=triangle*, mark size=1pt, c1, mark options={solid}, dotted] coordinates {(0, 0.463464) (1, 0.461448) (2, 0.44298399999999993) (3, 0.43684399999999995) (4, 0.4386880000000001) (5, 0.4406) (6, 0.440588)};
	\addplot[mark=pentagon*, mark size=1pt, c2, mark options={solid}] coordinates {(0, 0.6150800000000001) (1, 0.61804) (2, 0.61698) (3, 0.6225799999999999) (4, 0.60988) (5, 0.61878) (6, 0.6227)};
	\addplot[mark=pentagon*, mark size=1pt, c2, mark options={solid}, dotted] coordinates {(0, 0.48687199999999997) (1, 0.5156959999999999) (2, 0.533104) (3, 0.5312080000000001) (4, 0.5256080000000001) (5, 0.535728) (6, 0.5423560000000001)};
	\end{axis}
	\end{tikzpicture}%
~%
\begin{tikzpicture}
	\begin{axis}[hplot, ymin=0.4, ytick distance={0.1}, ymax=0.95, xtick=data, ymajorticks=false, xticklabels={10,25,50,100,250,500,1000},  title={\scriptsize\sffamily Steps ($n = 100$)}]
	\addplot[mark=*, mark size=1pt, c0, mark options={solid}] coordinates {(0, 0.8049473684210527) (1, 0.8650657894736843) (2, 0.874684210526316) (3, 0.8814736842105262) (4, 0.8824473684210525) (5, 0.8841052631578947) (6, 0.8833684210526316)};
	\addplot[mark=*, mark size=1pt, c0, mark options={solid}, dotted] coordinates {(0, 0.6678263157894737) (1, 0.7688684210526316) (2, 0.832678947368421) (3, 0.8620789473684211) (4, 0.8656894736842105) (5, 0.8717578947368422) (6, 0.8727842105263159)};
	\addplot[mark=triangle*, mark size=1pt, c1, mark options={solid}] coordinates {(0, 0.5208200000000001) (1, 0.53372) (2, 0.56494) (3, 0.5785) (4, 0.5692000000000002) (5, 0.56832) (6, 0.5641)};
	\addplot[mark=triangle*, mark size=1pt, c1, mark options={solid}, dotted] coordinates {(0, 0.43484) (1, 0.491476) (2, 0.528632) (3, 0.545572) (4, 0.5464960000000001) (5, 0.5445840000000001) (6, 0.540464)};
	\addplot[mark=pentagon*, mark size=1pt, c2, mark options={solid}] coordinates {(0, 0.6220800000000001) (1, 0.6411) (2, 0.6590999999999999) (3, 0.67676) (4, 0.68242) (5, 0.6841999999999999) (6, 0.68354)};
	\addplot[mark=pentagon*, mark size=1pt, c2, mark options={solid}, dotted] coordinates {(0, 0.47234) (1, 0.5618320000000001) (2, 0.606416) (3, 0.637436) (4, 0.6477999999999999) (5, 0.654888) (6, 0.65518)};
	\end{axis}
	\end{tikzpicture}%
	\vskip3pt
	
	\begin{tikzpicture}
	\begin{axis}[hplot,  ymin=0.4, ymax=0.95, ytick distance={0.1}, xtick=data, xticklabels={1,2,4,8,16,32},  title={\scriptsize\sffamily Batch size ($n = 10$)}]
	\addplot[mark=*, mark size=1pt, c0, mark options={solid}] coordinates {(0, 0.8160263157894736) (1, 0.8143947368421053) (2, 0.7941315789473684) (3, 0.8029473684210526) (4, 0.8025) (5, 0.8051578947368421)};
	\addplot[mark=*, mark size=1pt, c0, mark options={solid}, dotted] coordinates {(0, 0.7404578947368422) (1, 0.7391789473684212) (2, 0.7313105263157895) (3, 0.7418684210526316) (4, 0.7360052631578948) (5, 0.7377473684210526)};
	\addplot[mark=triangle*, mark size=1pt, c1, mark options={solid}] coordinates {(0, 0.45686) (1, 0.46068) (2, 0.4558999999999999) (3, 0.4615) (4, 0.46898) (5, 0.4694400000000001)};
	\addplot[mark=triangle*, mark size=1pt, c1, mark options={solid}, dotted] coordinates {(0, 0.44030800000000003) (1, 0.443044) (2, 0.43684399999999995) (3, 0.439236) (4, 0.441812) (5, 0.441144)};
	\addplot[mark=pentagon*, mark size=1pt, c2, mark options={solid}] coordinates {(0, 0.60592) (1, 0.6136600000000001) (2, 0.6225799999999999) (3, 0.60946) (4, 0.61476) (5, 0.60894)};
	\addplot[mark=pentagon*, mark size=1pt, c2, mark options={solid}, dotted] coordinates {(0, 0.53664) (1, 0.5333239999999998) (2, 0.5312080000000001) (3, 0.526868) (4, 0.524836) (5, 0.523896)};
	\end{axis}
	\end{tikzpicture}%
~%
\begin{tikzpicture}
	\begin{axis}[hplot, ymin=0.4, ymax=0.95, ytick distance={0.1}, xtick=data, ymajorticks=false, xticklabels={1,2,4,8,16,32},  title={\scriptsize\sffamily Batch size ($n = 100$)}]
	\addplot[mark=*, mark size=1pt, c0, mark options={solid}] coordinates {(0, 0.8574473684210526) (1, 0.8780789473684211) (2, 0.8814736842105262) (3, 0.8834736842105263) (4, 0.8855526315789474) (5, 0.8868421052631579)};
	\addplot[mark=*, mark size=1pt, c0, mark options={solid}, dotted] coordinates {(0, 0.8294894736842104) (1, 0.8561736842105263) (2, 0.8620789473684211) (3, 0.8612052631578948) (4, 0.8615842105263157) (5, 0.8634842105263157)};
	\addplot[mark=triangle*, mark size=1pt, c1, mark options={solid}] coordinates {(0, 0.53012) (1, 0.57162) (2, 0.5785) (3, 0.5749199999999999) (4, 0.5709000000000001) (5, 0.5728)};
	\addplot[mark=triangle*, mark size=1pt, c1, mark options={solid}, dotted] coordinates {(0, 0.49138800000000005) (1, 0.5342399999999999) (2, 0.545572) (3, 0.5466240000000001) (4, 0.54512) (5, 0.54528)};
	\addplot[mark=pentagon*, mark size=1pt, c2, mark options={solid}] coordinates {(0, 0.63358) (1, 0.66454) (2, 0.67676) (3, 0.6767799999999999) (4, 0.6767199999999999) (5, 0.6771999999999999)};
	\addplot[mark=pentagon*, mark size=1pt, c2, mark options={solid}, dotted] coordinates {(0, 0.6091599999999999) (1, 0.6289359999999999) (2, 0.637436) (3, 0.637776) (4, 0.637248) (5, 0.6414880000000001)};
	\end{axis}
	\end{tikzpicture}%
	
	\caption{Performance of \pet{} (solid lines) and average performance of individual models (dotted lines) for different learning rates (LR), training steps (Steps), and batch sizes. For reasons of readability, the legend is shown in the top left plot only.}
	\label{fig:performance-hyperparams}
\end{figure}

\question{Do we really need large amounts of unlabeled data?}
\label{q6}

One drawback of \pet{} compared to using individual PVPs is the additional need for unlabeled data; while such data is often available in real-world settings, there are cases in which even unlabeled examples are hard to obtain. As recent work shows that pretrained LMs are capable of generating data of reasonable quality themselves \citep{Anaby-Tavor_Carmeli_Goldbraich_Kantor_Kour_Shlomov_Tepper_Zwerdling_2020,papanikolaou2020dare,yang-etal-2020-generative,mohapatra2020simulated,kumar2021data,schick2021generating}, we investigate whether unlabeled data can simply be replaced by synthetic examples.

\paragraph{Setup}

We use GPT2-XL \citep{radford2018language} to generate synthetic unlabeled data. To this end, we provide one or two randomly chosen training examples without labels as \emph{in-context examples} \citep{brown2020language} and let the model generate one additional example. For two inputs $x_1$ and $x_2$, the input given to the model is
\[ \small\textsf{Example 1: } x_1 \textsf{ \return{}\, Example 2: } x_2 \textsf{ \return{}\, Example 3:} \]
where $\,\return{}\,$ denotes two consecutive line breaks. If an input consists of two texts, we simply concatenate them using the sequence \textsf{+++} as a separator. 

We generate 10,000 examples for $n = 10$ and 30,000 examples for $n = 100$ using
top-$p$ sampling \citep{Holtzman2020The} with $p = 0.9$. For each input, we stop the generation process as soon as the model generates two consecutive line breaks. We discard all examples for which the model does not generate two consecutive line breaks within 128 tokens; for datasets with text pairs, we also discard examples where the model fails to generate the sequence separator (\textsf{+++}).

As the datasets obtained with this method may be highly imbalanced regarding the distribution of (unknown) labels, we also experiment with a \emph{balanced} variant: We use the ensemble of models trained on individual PVPs to assign labels to each example and only keep so many examples per label that the resulting dataset -- which is used for training the final classifier -- is balanced.

\paragraph{Results} Figure~\ref{fig:performance-unlabeled-data-generation} shows the performance of individual patterns as well as \pet{} and \ipet{} both with real and synthetic unlabeled data. Except for \ipet{} on Yahoo Questions with $n = 10$, using synthetic data consistently achieves accuracies within one point of using real data, with our balanced version performing slightly better. Moreover, for $n = 10$ using synthetic data even improves accuracy in some cases. This shows that in the absence of unlabeled examples, synthetic data obtained from generative language models can serve as a drop-in replacement without substantially degrading performance.

\pgfplotsset{
	mbar/.style={
		title style={yshift=-1.5ex},
		ybar=2pt,
		ymajorgrids,
		ytick distance={20},
		major grid style={line width=.2pt,draw=decentgrey},
		symbolic x coords={10, 100},
		xticklabels={$n = 10$, $n = 100$}, 
		xtick=data,
		bar width=0.25cm,
		nodes near coords,
		nodes near coords align={vertical},
		axis line style={decentgrey!80!black},
		major tick style={decentgrey!80!black},
		xtick pos=left,
		ytick pos=left,
		ylabel near ticks,
		xlabel near ticks,
		tick align=outside,
		enlarge x limits=0.5,
		ymin=0,
		major tick length=0.075cm,
		width = 1.1\linewidth,
		height = 0.15\textheight,
		log ticks with fixed point,
		every node near coord/.append style={font=\sansmath\sffamily\tiny, inner sep=2pt, /pgf/number format/.cd, fixed, fixed zerofill, precision=0, /tikz/.cd},
		x tick label style={font=\sansmath\sffamily\tiny},
		y tick label style={font=\sansmath\sffamily\tiny, /pgf/number format/.cd, fixed, fixed zerofill, precision=0, /tikz/.cd},
	},
}

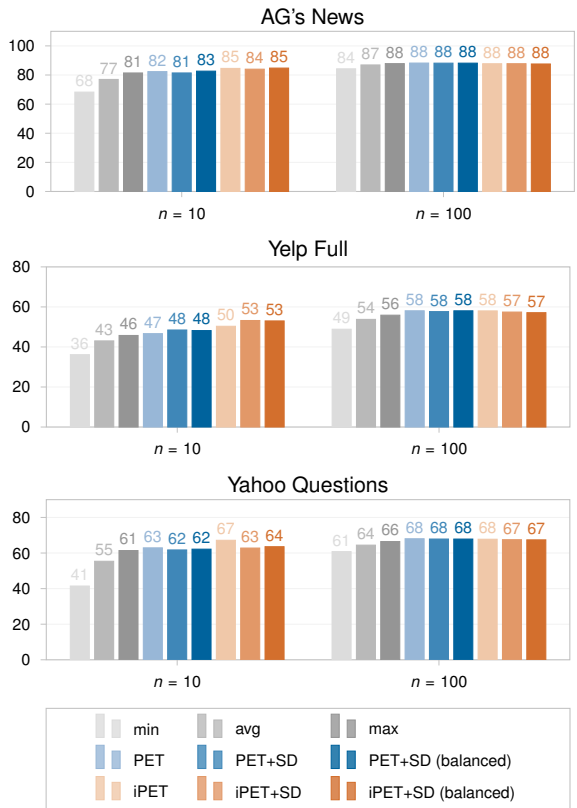
\begin{figure}
	\centering
	\begin{tikzpicture}
	\begin{axis}[mbar, title={\scriptsize\sffamily AG's News}, ymax=110]
	\addplot[darkgrey!33!white, fill = darkgrey!33!white] coordinates {(10,68.25263157894736) (100,84.2421052631579)};
	\addplot[darkgrey!66!white, fill = darkgrey!66!white] coordinates {(10,76.95389016018306) (100,86.90400457665903)};
	\addplot[darkgrey, fill = darkgrey] coordinates {(10,81.4) (100,87.81052631578949)};
	\addplot[c0!33!white, fill = c0!33!white] coordinates {(10,82.31315789473683) (100,88.20263157894736)};
	\addplot[c0!66!white, fill = c0!66!white] coordinates {(10,81.41052631578948) (100,88.13684210526317)};
	\addplot[c0, fill = c0] coordinates {(10,82.64736842105262) (100,88.13947368421053)};
	\addplot[c1!33!white, fill = c1!33!white] coordinates {(10,84.53157894736843) (100,87.73684210526315)};
	\addplot[c1!66!white, fill = c1!66!white] coordinates {(10,84.00000000000001) (100,87.75)};
	\addplot[c1, fill = c1] coordinates {(10,84.75263157894737) (100,87.57105263157895)};
	\end{axis}
	\end{tikzpicture}
	
	\begin{tikzpicture}
	\begin{axis}[mbar, title={\scriptsize\sffamily Yelp Full}, ymax=80]
	\addplot[darkgrey!33!white, fill = darkgrey!33!white] coordinates {(10,36.147999999999996) (100,48.84799999999999)};
	\addplot[darkgrey!66!white, fill = darkgrey!66!white] coordinates {(10,43.02504347826087) (100,53.767652173913035)};
	\addplot[darkgrey, fill = darkgrey] coordinates {(10,45.73600000000001) (100,55.874)};
	\addplot[c0!33!white, fill = c0!33!white] coordinates {(10,46.647999999999996) (100,58.06)};
	\addplot[c0!66!white, fill = c0!66!white] coordinates {(10,48.472) (100,57.66000000000001)};
	\addplot[c0, fill = c0] coordinates {(10,48.184000000000005) (100,58.05200000000001)};
	\addplot[c1!33!white, fill = c1!33!white] coordinates {(10,50.330000000000005) (100,58.004)};
	\addplot[c1!66!white, fill = c1!66!white] coordinates {(10,53.21) (100,57.428000000000004)};
	\addplot[c1, fill = c1] coordinates {(10,53.0) (100,57.118)};
	\end{axis}
	\end{tikzpicture}
	
	\begin{tikzpicture}
	\begin{axis}[mbar, title={\scriptsize\sffamily Yahoo Questions}, ymax=90, 	
	legend entries={min, avg, max, PET, PET+SD, PET+SD (balanced), iPET\phantom{xx}, iPET+SD\phantom{xx}, iPET+SD (balanced)},
	legend style={inner xsep=18.5pt, style={column sep=0.1cm}, draw=decentgrey!80!black, fill=white!75, text opacity =1, fill opacity=0.8, at={(0.5,-0.3)},anchor=north, font=\sffamily\tiny},
	legend cell align=left,
	legend columns=3,]
	\addplot[darkgrey!33!white, fill = darkgrey!33!white] coordinates {(10,41.454) (100,60.79199999999999)};
	\addplot[darkgrey!66!white, fill = darkgrey!66!white] coordinates {(10,55.3608695652174) (100,64.43278260869565)};
	\addplot[darkgrey, fill = darkgrey] coordinates {(10,61.388) (100,66.448)};
	\addplot[c0!33!white, fill = c0!33!white] coordinates {(10,62.91) (100,68.016)};
	\addplot[c0!66!white, fill = c0!66!white] coordinates {(10,61.734) (100,67.864)};
	\addplot[c0, fill = c0] coordinates {(10,62.168) (100,67.846)};
	\addplot[c1!33!white, fill = c1!33!white] coordinates {(10,67.16) (100,67.75)};
	\addplot[c1!66!white, fill = c1!66!white] coordinates {(10,62.798) (100,67.486)};
	\addplot[c1, fill = c1] coordinates {(10,63.556000000000004) (100,67.39999999999999)};
	\end{axis}
	\end{tikzpicture}
	\caption{Minimum, average and maximum performance of individual patterns compared to regular \pet{} and \ipet{} as well as \pet{} and \ipet{} with synthetic data (+SD). Accuracies are scaled by a factor of 100 for better readability.}
	\label{fig:performance-unlabeled-data-generation}
\end{figure}

\section{\pet{} for Real-World Tasks}
\label{sec:raft}

We use our insights from Section~\ref{sec:true-fsl} to apply \pet{} to the RAFT benchmark, a collection of 11 diverse real-world tasks whose automated solution has inherent value to someone \citep{alex2021raft}. These tasks pose various challenges to few-shot approaches as they require some amount of domain expertise and the ability to understand detailed instructions, to process long inputs and to handle a large number of output classes.

\paragraph{Tasks and Datasets} The RAFT benchmark includes 11 tasks from different domains: ADE Corpus V2 (ADE), Banking77 (B77), NeurIPS impact statement risks (NIS), OneStopEnglish (OSE), Overruling (Over), Semiconductor org types (SOT), Systematic review inclusion (SRI), TAI safety research (TAI), Terms of Service (ToS), TweetEval Hate (TEH) and Twitter complaints (TC). For a detailed overview of all tasks, we refer to \citet{alex2021raft}. Each task comes with 50 labeled training examples; in accordance with the RAFT rules, we additionally make use of the unlabeled data (ranging from 150 to 5,000 examples) for \pet{}'s distillation step. Note that the unlabeled set for RAFT is the same as the test set, meaning that unlike in Section~\ref{sec:true-fsl}, our final classifier is directly trained on (unlabeled) test examples.

\paragraph{PVPs} Based on our findings from Q\ref{q1} and Q\ref{q2}, we exclusively specify \textsc{Q\&A} prompts for each task. To obtain the question $q$, we make minimal changes to the original instructions of \citet{alex2021raft}; we rephrase all binary classification tasks as yes/no questions to facilitate finding a suitable verbalizer. For example, we rephrase the instruction ``Label the sentence based on whether it is related to an adverse drug effect (ADE).'' as the question ``Is this sentence related to an adverse drug effect (ADE)?'' 
Following our results from Q\ref{q4}, we specify 4 PVPs per task. In case of binary classification, we use two different patterns that either include or omit the full task specification of \citet{alex2021raft}\footnote{For a full list of all task specifications, see \url{https://github.com/oughtinc/raft-baselines}.} and combine them with both a yes/no verbalizer and a true/false verbalizer. The full set of PVPs for all tasks can be found in Appendix~\ref{appendix:raft-pvps}.

\paragraph{Hyperparameters} We mostly keep the default values for hyperparameters used throughout Section~\ref{sec:true-fsl} as our experiments for Q\ref{q5} show that these perform well for all tasks considered. However, we make the following changes: 
\begin{itemize}
	\item We replace RoBERTa (base) with ALBERT (xxlarge, v2). While being much slower to train, ALBERT was shown to outperform RoBERTa both in regular and few-shot settings \citep{lan2019albert,schick2020just,logan2021cutting}.
	\item For tasks with more than 4,000 unlabeled examples, we finetune the distilled model for 2,000 steps instead of 1,000 steps. Otherwise, some examples would not be seen at all during training.\footnote{Training for 1,000 steps with a batch size of 4 means that at most 4,000 examples can be processed.}
	\item Following \citet{schick2020exploiting,schick2020just} we train three individual models per PVP. This improves robustness as performance can vary largely between individual finetuning runs.
\end{itemize}  

\begin{table*}
	\small
	\begin{tabularx}{\linewidth}{lXXXXXXXXXXXX}
		\toprule
		\textbf{Method} & \textbf{ADE} & \textbf{B77} & \textbf{NIS} & \textbf{OSE} & \textbf{Over} & \textbf{SOT} & \textbf{SRI} & \textbf{TAI} & \textbf{ToS} & \textbf{TEH} & \textbf{TC}  & \textbf{Avg} \\
		\midrule
		GPT-2 & 60.0 & 12.1 & 56.1 & 24.5 & 49.8 & 38.0 & 49.2 & 61.2 & 49.8 & 31.1 & 72.3 & 45.8 \\
		GPT-Neo & 45.2 & 14.9 & 40.8 & 34.3 & 68.1 & 40.6 & 49.3 & 60.5 & 56.5 & 55.4 & 63.6 & 48.1  \\
		AdaBoost & 54.3 & 02.3 & 62.6 & 47.5 & 83.8 & 45.5 & 50.6 & 55.6 & 56.0 & 44.3 & 62.5 & 51.4 \\
		snlt & 60.3 & 24.8 & 58.5 & 30.2 & 83.1 & 33.6 & 49.2 & 62.6 & 54.0 & 44.9 & 79.1 & 52.8 \\
		GPT-3 & 68.6 & 29.9 & 67.9 & 43.1 & \bt\underline{93.7} & 76.9 & \bt\underline{51.6} & \bt\underline{65.6} & 57.4 & 52.6 & 82.1 & 62.7 \\
		SetFit & 72.6 & 53.8 & \bt\underline{87.2} & 52.1 & 90.7 & 68.2 & 49.3 & 62.8 & \bt 62.0 & \bt 53.2 & \bt 83.7 & 66.9 \\
		\pet{} & \bt 82.2 & \bt 59.3 & 85.7 & \bt\underline{64.6} & 90.8 & \bt 81.6 & 49.3 & 63.8 & 57.6 & 48.3 & 82.4 & \bt 69.6 \\
		\midrule
		Human & \underline{83.0} & \underline{60.7} & 85.7 & \underline{64.6} & 91.7 & \underline{90.8} & 46.8 & 60.9 & \underline{62.7} & \underline{72.2} & \underline{89.7} & \underline{73.5} \\
		\bottomrule
	\end{tabularx}
	\caption{Performance of various baselines and \pet{} on the RAFT benchmark \citep{alex2021raft}; shown numbers are macro F1 scores multiplied by 100. Best model performance is shown in bold, best overall performance (including human annotators) is underlined. The final column shows average performance across all 11 tasks.}
	\label{table:performance-raft}
\end{table*}

\paragraph{Handling Many Labels} The B77 dataset contains online banking customer service queries annotated with one of 77 possible intents. This large amount of different outputs leads to several issues for \pet{}: First, it is impossible to specify a meaningful verbalizer that maps each intent to a single token. We initially experimented with the multi-mask version of \pet{} \citep{schick2020just}, but found it to be too inefficient to get results in a reasonable amount of time. Therefore, we tried the following solution: We rephrase the task as binary classification, where for each pair of query $x$ and intent $y$, the task is to decide whether $y$ is the correct intent for $x$. For each original training example $(x,y)$, we generate one example $(x,y,\textsf{\small True})$ and four examples $(x,y', \textsf{\small False})$ with randomly sampled, wrong intents $y'$. As this increases the amount of data fivefold, we finetune each individual model for 500 steps instead of 100 steps. 

While this approach solves our problem, it is still not particularly efficient: Reframing the task as binary classification means that for each input, 77 forward passes are required to find the correct intent. We thus train the final model as a regular classifier with 77 different output classes; for training this classifier on an input $x$, we set the target probability of each output $y$ proportional to the probability of $\textsf{\small True}$ being the correct output for $(x, y)$ according to our ensemble of binary classifiers.

Finally, another issue is that with 50 labeled examples, at least 27 labels are not covered in the training set at all; this may bias a model to never predict any of these labels. To alleviate this issue, we train two generations of models using \ipet{}. For training the second generation, we obtain training data covering all possible labels as follows: For each label, we pick the two examples from our set of unlabeled data for which this label is most likely according to the first generation; the same approach was used by \citet{schick2020exploiting} for applying \ipet{} in zero-shot settings.

Of course, the nature of RAFT makes it impossible to measure the impact of any of these choices. While we could conduct experiments similar to those in Section~\ref{sec:true-fsl}, none of the datasets considered therein has a similar structure to B77; as our modifications affect only one out of 11 tasks, we thus decided to not perform any further analysis. 

\paragraph{Monitoring} We checked for \textsc{Train Set Underfitting} and \textsc{Constant Predictions} as in Section~\ref{sec:true-fsl} to detect finetuning issues. Unlike for our experiments in Section~\ref{sec:true-fsl}, on RAFT we encountered some issues that could \emph{not} be resolved simply by retraining with a different seed:
\begin{itemize}
\item We observed \textsc{Train Set Underfitting} for the final classifier on B77. This may be due to the classification head for 77 classes introducing many new parameters; we tried training the final model for 5,000 steps instead of 2,000 steps, which fixed this issue.

\item We observed \textsc{Constant Predictions} for the ToS training set. Doubling the number of training steps resolved this problem. 

\item Finally, we also observed \textsc{Constant Predictions} on the unlabeled data of SRI. Upon manually inspecting the training set, we observed that all but one out of 50 examples have the same label. As all models already classified the training set perfectly, we left the setup for our SRI submission unchanged.
\end{itemize}

\paragraph{Results} For all 11 tasks, results of \pet{} and various baselines are shown in Table~\ref{table:performance-raft}.\footnote{\lsstyle All results are taken directly from the leaderboard at \url{https://huggingface.co/spaces/ought/raft-leaderboard}.} As can be seen, \pet{} performs better than all other approaches on average, achieving near-human performance for 7 out of 11 tasks. Note however that non-expert humans perform worse than a majority baseline SRI, so results on this task should be taken with a grain of salt. \pet{} also clearly outperforms a GPT-3 model \citep{brown2020language} by almost 7 points, despite the latter being larger by several orders of magnitude.\footnote{We were unable to find any information on SetFit, the second-best performing method (e.g., which underlying LM is used or whether it makes use of any additional data), so we cannot make a fair comparison.} While \pet{} is particularly successful on ADE, B77 and OSE (where it outperforms GPT-3 by 13.6, 21.5 and 29.4 points, respectively), it performs comparably bad on datasets in the law (Over, ToS) and social media (TEH, TC) domain. Our approach for handling many labels performs surprisingly well on B77 without any tuning of its parameters. Due to the nature of the RAFT benchmark, we cannot perform further analysis or ablation studies.

\section{Discussion}

Our experimental results in Section~\ref{sec:true-fsl}~and~\ref{sec:raft} show that strong performance in few-shot settings is clearly possible without manual prompt tuning or hyperparameter optimization on large development sets; in other words, \pet{} can successfully be applied in true few-shot settings. While we believe that it should be an important goal of future work to make LMs more robust to different instructions, even with current models it is relatively easy to successfully apply \pet{} when following a few simple principles -- such as rephrasing the task in a Q\&A format, using simple vocabulary and single-token verbalizers where possible, and specifying at least a handful of different patterns.
In light of these findings, we also hope that future work will not view human involvement in prompt design as a drawback of instruction-based approaches, but rather as an exciting possibility to communicate with models in ways other than exclusively through examples.

There are various limitations to our study. First, a major obstacle to actually applying \pet{} in real-world applications is that we do not know a priori how well it performs for a given task; we therefore believe an important next step is to investigate methods for estimating performance without access to large test sets -- for example, through model calibration \citep{desai2020calibration,jiang2021know} -- in real-world settings. In addition, we did not fully explore the capabilities of \pet{}; for example, we did not investigate domain-adaptive pretraining \citep{gururangan-etal-2020-dont} and auxiliary language modeling \citep{chronopoulou-etal-2019-embarrassingly}, both of which were shown to be helpful by \citet{schick2020exploiting}. We also did not quantify the impact of our decisions regarding B77 and the effectiveness of our monitoring and only considered English models and datasets. Finally, we did not examine \pet{}'s performance beyond aggregate scores. While this is not feasible on RAFT due to the nature of this dataset, performing such analysis either with other datasets or with methods such as the ones proposed by \citet{ribeiro2020accuracy} would be relevant future work to understand real-world capabilities of instruction-based approaches more comprehensively.

\section{Conclusion}

In light of recent work casting doubt on the performance of prompt-based approaches in true few-shot settings \citep{perez2021true}, we have conducted an extensive study of \pet{}. In a controlled environment, we found that manually designed instructions consistently outperform null prompts, with \textsc{Q\&A}-style prompts performing best (Q\ref{q1}, Q\ref{q2}). Across different tasks, models and training set sizes, \pet{} consistently outperforms even the best individual prompt (Q\ref{q1}, Q\ref{q2}). We have also shown that \pet{} is robust to uninformative prompts and to different choices of hyperparameters (Q\ref{q3}, Q\ref{q5}), that as little as four prompts are sufficient to reach good performance (Q\ref{q4}), and that synthetic examples can be used to replace large amounts of unlabeled data (Q\ref{q6}). On the basis of these insights, we applied \pet{} to a benchmark of real-world tasks, where it achieves near-human performance for 7 out of 11 tasks without any tuning on a development set, demonstrating the power of instruction-based approaches in true few-shot settings.

\bibliography{literature}
\bibliographystyle{acl_natbib}

\appendix
\section{PVPs for AG's, Yelp, Yahoo}
\label{appendix:default-pvps}

Below, we list the 23 patterns and the single verbalizer used for each task throughout Section~\ref{sec:true-fsl}. We group all patterns by their category.

\subsection{AG's News} 

\paragraph{Inputs}
\begin{itemize}
	\item $x_1$: The headline of the news article to be classified.
	\item $x_2$: The text body of the news article to be classified.
\end{itemize}

\paragraph{\textsc{Null} Patterns}
\begin{itemize}
	\item \textsf{[MASK]} $x_1$ $x_2$
	\item $x_1$ $x_2$ \textsf{[MASK]}
	\item $x_1$ \textsf{[MASK]} $x_2$
\end{itemize}

\paragraph{\textsc{Punc} Patterns}
\begin{itemize}
	\item \textsf{[MASK] : $x_1$ $x_2$}
	\item \textsf{[MASK] - $x_1$ $x_2$}
	\item \textsf{[MASK] . $x_1$ $x_2$}
	\item \textsf{( [MASK] ) $x_1$ $x_2$}	
	\item \textsf{$x_1$ ( [MASK] ) $x_2$}	
	\item \textsf{$x_1$ $x_2$ ( [MASK] )}
\end{itemize}

\paragraph{\textsc{Prompt} Patterns}
\begin{itemize}
	\item \textsf{$x_1$ $x_2$ Category: [MASK].}
	\item \textsf{$x_1$ $x_2$ Class: [MASK].}
	\item \textsf{$x_1$ $x_2$ Topic: [MASK].}
	\item \textsf{$x_1$ $x_2$ Theme: [MASK].}
	\item \textsf{$x_1$ $x_2$ Category: [MASK]}
	\item \textsf{$x_1$ $x_2$ Class: [MASK]}
	\item \textsf{$x_1$ $x_2$ Topic: [MASK]}
	\item \textsf{$x_1$ $x_2$ Theme: [MASK]}
	\item \textsf{[MASK] News: $x_1$ $x_2$}
	\item \textsf{[MASK] NEWS: $x_1$ $x_2$}
\end{itemize}

\paragraph{\textsc{Q\&A} Patterns}
\begin{itemize}
		\item \textsf{$x_1$ $x_2$ Question: What is the topic of this article? Answer: [MASK].}		
		\item \textsf{$x_1$ $x_2$ Question: What is the category of this article? Answer: [MASK].}		
		\item \textsf{$x_1$ $x_2$ Question: What is the topic of this article? Answer: [MASK]}		
		\item \textsf{$x_1$ $x_2$ Question: What is the category of this article? Answer: [MASK]}
\end{itemize}

\paragraph{Verbalizer} \textsf{World $\mapsto$ World, Sports $\mapsto$ Sports, Business $\mapsto$ Business, Science/Tech $\mapsto$ Tech}
  
\subsection{Yelp Reviews Full Star} 

\paragraph{Inputs}
\begin{itemize}
	\item $x$: The review to be classified.
\end{itemize}

\paragraph{\textsc{Null} Patterns}
\begin{itemize}
	\item \textsf{[MASK]} $x$
	\item $x$ \textsf{[MASK]}
\end{itemize}

\paragraph{\textsc{Punc} Patterns}
\begin{itemize}
	\item \textsf{[MASK] : $x$}
	\item \textsf{[MASK] - $x$}
	\item \textsf{[MASK] . $x$}
	\item \textsf{( [MASK] ) $x$}		
	\item \textsf{$x$ ( [MASK] )}
\end{itemize}

\paragraph{\textsc{Prompt} Patterns}
\begin{itemize}
	\item \textsf{$x$ It was [MASK]}
	\item \textsf{$x$ All in all, it was [MASK]}
	\item \textsf{$x$ In summary, it was [MASK]}
	\item \textsf{$x$ It was [MASK].}
	\item \textsf{$x$ All in all, it was [MASK].}
	\item \textsf{$x$ In summary, it was [MASK].}
	\item \textsf{$x$ The restaurant was [MASK]}
	\item \textsf{$x$ All in all, the restaurant was [MASK]}
	\item \textsf{$x$ In summary, the restaurant was [MASK]}
	\item \textsf{$x$ The restaurant was [MASK].}
	\item \textsf{$x$ All in all, the restaurant was [MASK].}
	\item \textsf{$x$ In summary, the restaurant was [MASK].}
\end{itemize}

\paragraph{\textsc{Q\&A} Patterns}
\begin{itemize}
	\item \textsf{$x$ Question: What does the customer think of this restaurant? Answer: It is [MASK].}		
	\item \textsf{$x$ Question: What does the customer think of this place? Answer: It is [MASK].}		
	\item \textsf{$x$ Question: What does the customer think of this restaurant? Answer: It is [MASK]}		
	\item \textsf{$x$ Question: What does the customer think of this place? Answer: It is [MASK]}
\end{itemize}

\paragraph{Verbalizer} \textsf{1 $\mapsto$ terrible, 2 $\mapsto$ bad, 3 $\mapsto$ okay, 4 $\mapsto$ good, 5 $\mapsto$ great}

\subsection{Yahoo Questions} 

\paragraph{Inputs}
\begin{itemize}
	\item $x_1$: The question to be classified.
	\item $x_2$: The top answer to the question.
\end{itemize}

\paragraph{\textsc{Null} Patterns}
\begin{itemize}
	\item \textsf{[MASK]} $x_1$ $x_2$
	\item $x_1$ $x_2$ \textsf{[MASK]}
	\item $x_1$ \textsf{[MASK]} $x_2$
\end{itemize}

\paragraph{\textsc{Punc} Patterns}
\begin{itemize}
	\item \textsf{[MASK] : $x_1$ $x_2$}
	\item \textsf{[MASK] - $x_1$ $x_2$}
	\item \textsf{[MASK] . $x_1$ $x_2$}
	\item \textsf{( [MASK] ) $x_1$ $x_2$}	
	\item \textsf{$x_1$ ( [MASK] ) $x_2$}	
	\item \textsf{$x_1$ $x_2$ ( [MASK] )}
\end{itemize}

\paragraph{\textsc{Prompt} Patterns}
\begin{itemize}
	\item \textsf{$x_1$ $x_2$ Category: [MASK].}
	\item \textsf{$x_1$ $x_2$ Class: [MASK].}
	\item \textsf{$x_1$ $x_2$ Topic: [MASK].}
	\item \textsf{$x_1$ $x_2$ Theme: [MASK].}
	\item \textsf{$x_1$ $x_2$ Category: [MASK]}
	\item \textsf{$x_1$ $x_2$ Class: [MASK]}
	\item \textsf{$x_1$ $x_2$ Topic: [MASK]}
	\item \textsf{$x_1$ $x_2$ Theme: [MASK]}
	\item \textsf{[MASK] Question: $x_1$ $x_2$}
	\item \textsf{[MASK] QUESTION: $x_1$ $x_2$}
\end{itemize}

\paragraph{\textsc{Q\&A} Patterns}
\begin{itemize}
	\item \textsf{$x_1$ $x_2$ Question: What is the topic of this question? Answer: [MASK].}		
	\item \textsf{$x_1$ $x_2$ Question: What is the category of this question? Answer: [MASK].}		
	\item \textsf{$x_1$ $x_2$ Question: What is the topic of this question? Answer: [MASK]}		
	\item \textsf{$x_1$ $x_2$ Question: What is the category of this question? Answer: [MASK]}
\end{itemize}

\paragraph{Verbalizer} \textsf{Society \& Culture $\mapsto$ Society, Science \& Mathematics $\mapsto$ Science, Health $\mapsto$ Health, Education \& Reference $\mapsto$ Education, Computers \& Internet $\mapsto$ Computer, Sports $\mapsto$ Sports, Business \& Finance $\mapsto$ Business, Entertainment \& Music $\mapsto$ Entertainment, Family \& Relationships $\mapsto$ Relationship, Politics \& Government $\mapsto$ Politics}

\section{PVPs for RAFT}
\label{appendix:raft-pvps}

Below, we list the PVPs used for all tasks in the RAFT benchmark. For each task $t$, we make use of the original task description $D_t$ that we copy verbatim from \citet{alex2021raft}.\footnote{See \url{https://github.com/oughtinc/raft-baselines/tree/master/example_prompts}.} We use two vertical bars (||) to mark
boundaries between text segments. If multiple patterns and verbalizers are specified, each verbalizer is used in combination with each pattern.

\subsection{ADE}

\paragraph{Task Description} \textsf{Label the sentence based on whether it is related to an adverse drug effect (ADE). Details are described below:\\
	Drugs: Names of drugs and chemicals that include brand names, trivial names, abbreviations and systematic names were annotated. Mentions of drugs or chemicals should strictly be in a therapeutic context. This category does not include the names of metabolites, reaction byproducts, or hospital chemicals (e.g. surgical equipment disinfectants).\\
	Adverse effect: Mentions of adverse effects include signs, symptoms, diseases, disorders, acquired abnormalities, deficiencies, organ damage or death that strictly occur as a consequence of drug intake.}

\paragraph{Inputs}
\begin{itemize}
	\item $x$: The text to be classified.
\end{itemize}

\paragraph{Patterns}
\begin{itemize}
	\item $D_t$ ||  $x$ \textsf{Question: Is this sentence related to an adverse drug effect (ADE)? Answer: [MASK].}
	\item $x$ \textsf{Question: Is this sentence related to an adverse drug effect (ADE)? Answer: [MASK].}
\end{itemize}

\paragraph{Verbalizers}
\begin{itemize}
\item \textsf{not ADE-related $\mapsto$ No, ADE-related $\mapsto$ Yes}
\item \textsf{not ADE-related $\mapsto$ False, ADE-related $\mapsto$ True}
\end{itemize}

\subsection{B77}

\paragraph{Task Description} \textsf{The following is a banking customer service query. Classify the query into one of the 77 categories available.}

\paragraph{Inputs}
\begin{itemize}
	\item $x$: The text to be classified.
	\item $y$: The correct intent for the given text.
\end{itemize}

\paragraph{Patterns}
\begin{itemize}
	\item $D_t$ ||  $x$ \textsf{Question: Is $y$ the correct category for this query? Answer: [MASK].}
	\item $x$ \textsf{Question: Is $y$ the correct category for this query? Answer: [MASK].}
\end{itemize}

\paragraph{Verbalizers}
\begin{itemize}
	\item \textsf{False $\mapsto$ No, True $\mapsto$ Yes}
	\item \textsf{False $\mapsto$ False, True $\mapsto$ True}
\end{itemize}

\subsection{NIS}

\paragraph{Task Description} \textsf{Label the impact statement based on whether it mentions a harmful application of the research done in the paper. Make sure the statement is sufficient to conclude there are harmful applications of the research being done, not a past risk that this research is solving.}

\paragraph{Inputs}
\begin{itemize}
	\item $x$: The text to be classified.
\end{itemize}

\paragraph{Patterns}
\begin{itemize}
	\item $D_t$ ||  $x$ \textsf{Question: Does this impact statement mention a harmful application? Answer: [MASK].}
	\item $x$ \textsf{Question: Does this impact statement mention a harmful application? Answer: [MASK].}
\end{itemize}

\paragraph{Verbalizers}
\begin{itemize}
	\item \textsf{doesn't mention a harmful application $\mapsto$ No, mentions a harmful application $\mapsto$ Yes}
	\item \textsf{doesn't mention a harmful application $\mapsto$ False, mentions a harmful application $\mapsto$ True}
\end{itemize}

\subsection{OSE}

\paragraph{Task Description} \textsf{The following is an article sourced from The Guardian newspaper, and rewritten by teachers to suit three levels of adult English as Second Language (ESL) learners: elementary, intermediate, and advanced. Predict the level of the article.}

\paragraph{Inputs}
\begin{itemize}
	\item $x$: The text to be classified.
\end{itemize}

\paragraph{Patterns}
\begin{itemize}
	\item $D_t$ ||  $x$ \textsf{Question: What is the level of this article? Answer: [MASK].}
	\item $x$ \textsf{Question: What is the level of this article? Answer: [MASK].}
	\item $D_t$ ||  $x$ \textsf{Question: Is the level of this article ``elementary'', ``intermediate'' or ``advanced''? Answer: [MASK].}
	\item $x$ \textsf{Question: Is the level of this article ``elementary'', ``intermediate'' or ``advanced''? Answer: [MASK].}
\end{itemize}

\paragraph{Verbalizers}
\begin{itemize}
	\item \textsf{elementary $\mapsto$ elementary, intermediate $\mapsto$ intermediate, advanced $\mapsto$ advanced}
\end{itemize}

\subsection{Over}

\paragraph{Task Description} \textsf{In law, an overruling sentence is a statement that nullifies a previous case decision as a precedent, by a constitutionally valid statute or a decision by the same or higher ranking court which establishes a different rule on the point of law involved. Label the sentence based on whether it is overruling or not.}

\paragraph{Inputs}
\begin{itemize}
	\item $x$: The text to be classified.
\end{itemize}

\paragraph{Patterns}
\begin{itemize}
	\item $D_t$ ||  $x$ \textsf{Question: Is this sentence overruling? Answer: [MASK].}
	\item $x$ \textsf{Question: Is this sentence overruling? Answer: [MASK].}
\end{itemize}

\paragraph{Verbalizers}
\begin{itemize}
	\item \textsf{not overruling $\mapsto$ No, overruling $\mapsto$ Yes}
	\item \textsf{not overruling $\mapsto$ False, overruling $\mapsto$ True}
\end{itemize} 

\subsection{SOT}

\paragraph{Task Description} \textsf{The dataset is a list of institutions that have contributed papers to semiconductor conferences in the last 25 years, as catalogued by IEEE and sampled randomly. The goal is to classify the institutions into one of three categories: "university", "company" or "research institute".}

\paragraph{Inputs}
\begin{itemize}
	\item $x_1$: The title of the paper to be classified.
	\item $x_2$: The name of the organization to be classified.
\end{itemize}

\paragraph{Patterns}
\begin{itemize}
	\item $D_t$ || \textsf{Organization name:} $x_1$ \textsf{Paper title:} $x_2$ \textsf{Question: What is the category of this institution? Answer: [MASK].}
	\item \textsf{Organization name:} $x_1$ \textsf{Paper title:} $x_2$ \textsf{Question: What is the category of this institution? Answer: [MASK].}
	\item $D_t$ || \textsf{Paper title:} $x_2$  \textsf{Organization name:} $x_1$ \textsf{Question: What is the category of this institution? Answer: [MASK].}
	\item \textsf{Paper title:} $x_2$  \textsf{Organization name:} $x_1$ \textsf{Question: What is the category of this institution? Answer: [MASK].}

\end{itemize}

\paragraph{Verbalizers}
\begin{itemize}
	\item \textsf{company $\mapsto$ company, research institute $\mapsto$ institute, university $\mapsto$ university}
\end{itemize}

\subsection{SRI}

\paragraph{Task Description} \textsf{Identify whether this paper should be included in a meta-review which includes the findings of systematic reviews on interventions designed to promote charitable donations.\\
Included reviews should describe monetary charitable donations, assess any population of participants in any context, and be peer reviewed and written in English.\\
They should not report new data, be non-systematic reviews, consider cause-related marketing or other kinds of prosocial behaviour.}

\paragraph{Inputs}
\begin{itemize}
	\item $x_1$: The title of the paper to be classified.	
	\item $x_2$: The abstract of the paper to be classified.
	\item $x_3$: The journal of the paper to be classified.
\end{itemize}

\paragraph{Patterns}
\begin{itemize}
	\item $D_t$ ||  \textsf{Title:} $x_1$ \textsf{ Abstract:} $x_2$ \textsf{Journal:} $x_3$ \textsf{Question: Should this paper be included in a meta-review which includes the findings of systematic reviews on interventions designed to promote charitable donations? Answer: [MASK].}
	\item  \textsf{Title:} $x_1$ \textsf{ Abstract:} $x_2$ \textsf{Journal:} $x_3$ \textsf{Question: Should this paper be included in a meta-review which includes the findings of systematic reviews on interventions designed to promote charitable donations? Answer: [MASK].}
\end{itemize}

\paragraph{Verbalizers}
\begin{itemize}
	\item \textsf{not included $\mapsto$ No, included $\mapsto$ Yes}
	\item \textsf{not included $\mapsto$ False, included $\mapsto$ True}
\end{itemize} 

\subsection{TAI}

\paragraph{Task Description} \textsf{Transformative AI (TAI) is defined as AI that precipitates a transition comparable to (or more significant than) the agricultural or industrial revolution. Label a paper as "TAI safety research" if: \\
	1. The contents of the paper are directly motivated by, and substantively inform, the challenge of ensuring good outcomes for TAI, \\
	2. There is substantive content on AI safety, not just AI capabilities, \\
	3. The intended audience is the community of researchers, \\
	4. It meets a subjective threshold of seriousness/quality, \\
	5. Peer review is not required.}

\paragraph{Inputs}
\begin{itemize}
	\item $x_1$: The title of the paper to be classified.	
	\item $x_2$: The abstract of the paper to be classified.
\end{itemize}

\paragraph{Patterns}
\begin{itemize}
	\item $D_t$ ||  \textsf{Title:} $x_1$ \textsf{ Abstract:} $x_2$ \textsf{Question: Is this paper a TAI safety research paper? Answer: [MASK].}
	\item  \textsf{Title:} $x_1$ \textsf{ Abstract:} $x_2$ \textsf{Question: Is this paper a TAI safety research paper? Answer: [MASK].}
\end{itemize}

\paragraph{Verbalizers}
\begin{itemize}
	\item \textsf{not TAI safety research $\mapsto$ No, TAI safety research $\mapsto$ Yes}
	\item \textsf{not TAI safety research $\mapsto$ False, TAI safety research $\mapsto$ True}
\end{itemize} 

\subsection{ToS}

\paragraph{Task Description} \textsf{Label the sentence from a Terms of Service based on whether it is potentially unfair. If it seems clearly unfair, mark it as potentially unfair.\\
	According to art. 3 of the Directive 93/13 on Unfair Terms in Consumer Contracts, a contractual term is unfair if: 1) it has not been individually negotiated; and 2) contrary to the requirement of good faith, it causes a significant imbalance in the parties rights and obligations, to the detriment of the consumer.}

\paragraph{Inputs}
\begin{itemize}
	\item $x$: The text to be classified.	
\end{itemize}

\paragraph{Patterns}
\begin{itemize}
	\item $D_t$ || $x$ \textsf{Question: Is this sentence potentially unfair? Answer: [MASK].}
	\item  $x$ \textsf{Question: Is this sentence potentially unfair? Answer: [MASK].}
\end{itemize}

\paragraph{Verbalizers}
\begin{itemize}
	\item \textsf{not potentially unfair $\mapsto$ No, potentially unfair $\mapsto$ Yes}
	\item \textsf{not potentially unfair $\mapsto$ False, potentially unfair $\mapsto$ True}
\end{itemize} 

\subsection{TEH}

\paragraph{Task Description} \textsf{Label whether the following tweet contains hate speech against either immigrants or women. Hate Speech (HS) is commonly defined as any communication that disparages a person or a group on the basis of some characteristic such as race, color, ethnicity, gender, sexual orientation, nationality, religion, or other characteristics.}

\paragraph{Inputs}
\begin{itemize}
	\item $x$: The text to be classified.	
\end{itemize}

\paragraph{Patterns}
\begin{itemize}
	\item $D_t$ || $x$ \textsf{Question: Does this tweet contain hate speech against either immigrants or women? Answer: [MASK].}
	\item  $x$ \textsf{Question: Does this tweet contain hate speech against either immigrants or women? Answer: [MASK].}
\end{itemize}

\paragraph{Verbalizers}
\begin{itemize}
	\item \textsf{not hate speech $\mapsto$ No, hate speech $\mapsto$ Yes}
	\item \textsf{not hate speech $\mapsto$ False, hate speech $\mapsto$ True}
\end{itemize} 

\subsection{TC}

\paragraph{Task Description} \textsf{A complaint presents a state of affairs which breaches the writer’s favorable expectation. Label the tweet text based on whether it contains a complaint.}

\paragraph{Inputs}
\begin{itemize}
	\item $x$: The text to be classified.	
\end{itemize}

\paragraph{Patterns}
\begin{itemize}
	\item $D_t$ || $x$ \textsf{Question: Does this tweet text contain a complaint? Answer: [MASK].}
	\item  $x$ \textsf{Question: Does this tweet text contain a complaint? Answer: [MASK].}
\end{itemize}

\paragraph{Verbalizers}
\begin{itemize}
	\item \textsf{no complaint $\mapsto$ No, complaint $\mapsto$ Yes}
	\item \textsf{no complaint $\mapsto$ False, complaint $\mapsto$ True}
\end{itemize} 

\end{document}